\documentclass{article}

\PassOptionsToPackage{numbers,sort&compress}{natbib}

\usepackage[preprint]{neurips_2024}

\usepackage[utf8]{inputenc} 
\usepackage[T1]{fontenc}    
\usepackage{url}            
\usepackage{booktabs}       
\usepackage{amsfonts}       
\usepackage{graphicx}       
\usepackage{nicefrac}       
\usepackage{microtype}      
\usepackage{placeins}
\usepackage[most]{tcolorbox}
\usepackage{float} 
\usepackage{csvsimple}
\usepackage{pgfplotstable}
\pgfplotsset{compat=1.18}

\usepackage{hyperref}       
\usepackage{colortbl}
\usepackage{xcolor} 

\usepackage{caption}    

\usepackage[utf8]{inputenc}
\usepackage{booktabs} 

\usepackage{tikz}
\usetikzlibrary{matrix,calc}

\usepackage{soul}     
\sethlcolor{yellow}   

\usepackage{listings}

\usepackage{seqsplit}

\usepackage{threeparttable}
\usepackage{enumitem}

\usepackage{makecell}

\newenvironment{paddedtable}[1][ht]{%
  \setlength{\tabcolsep}{6pt}
  \begin{table}[#1]
}{%
  \end{table}
}

\definecolor{myblue}{HTML}{B3D9FF}   
\definecolor{myteal}{HTML}{B0C4CE}   
\definecolor{myyellow}{HTML}{FFFFAA} 
\definecolor{mypink}{HTML}{FFB3B3}   
\definecolor{myorange}{HTML}{FFD59A} 
\definecolor{mygreen}{HTML}{C3E6C3}  

\definecolor{warmgreen}{HTML}{659287}
\definecolor{burntorange}{HTML}{DEAA79}

\definecolor{myLightGray}{gray}{0.88}

\newtcolorbox{promptbox}{
  colback=warmgreen!15,  
  colframe=warmgreen,    
  title=Prompt,
  fonttitle=\bfseries,
  boxrule=1pt,
  left=2mm, right=2mm, top=2mm, bottom=2mm
}

\newtcolorbox{answerbox}{
  colback=burntorange!15,
  colframe=burntorange,
  title=Expected Answer,
  fonttitle=\bfseries,
  boxrule=1pt,
  left=2mm, right=2mm, top=2mm, bottom=2mm
}

\newtcolorbox{modelanswerbox}{
  colback=burntorange!15,
  colframe=burntorange,
  title=Model Answer,
  fonttitle=\bfseries,
  boxrule=1pt,
  left=2mm, right=2mm, top=2mm, bottom=2mm
}

\title{Assessing the Chemical Intelligence of Large Language Models}

\author{
  Nicholas T. Runcie \\
  Department of Statistics\\
  University of Oxford\\
  Oxford, UK\\
  \texttt{runcie@stats.ox.ac.uk} \\
  \And
  Charlotte M. Deane \\
  Department of Statistics\\
  University of Oxford\\
  Oxford, UK\\
  \texttt{deane@stats.ox.ac.uk} \\
  \And
  Fergus Imrie \\
  Department of Statistics\\
  University of Oxford\\
  Oxford, UK\\
  \texttt{imrie@stats.ox.ac.uk} \\
}

\begin{document}

\maketitle

\begin{abstract}
Large Language Models are versatile, general-purpose tools with a wide range of applications. Recently, the advent of ``reasoning models'' has led to substantial improvements in their abilities in advanced problem-solving domains such as mathematics and software engineering. In this work, we assessed the ability of reasoning models to perform chemistry tasks directly, without any assistance from external tools.
We created a novel benchmark, called ChemIQ, consisting of 816 questions assessing core concepts in organic chemistry, focused on molecular comprehension and chemical reasoning. 
Unlike previous benchmarks, which primarily use multiple choice formats, our approach requires models to construct short-answer responses, more closely reflecting real-world applications.
The reasoning models, OpenAI's o3-mini, Google's Gemini Pro 2.5, and DeepSeek R1, answered 50\%-57\% of questions correctly in the highest reasoning modes, with higher reasoning levels significantly increasing performance on all tasks. These models substantially outperformed the non-reasoning models which achieved only 3\%-7\% accuracy.
We found that Large Language Models can now convert SMILES strings to IUPAC names, a task earlier models were unable to perform. 
Additionally, we show that the latest reasoning models can elucidate structures from 1D and 2D $^{1}$H and $^{13}$C NMR data, with Gemini Pro 2.5 correctly generating SMILES strings for around 90\% of molecules containing up to 10 heavy atoms, and in one case solving a structure comprising 25 heavy atoms. 
For each task, we found evidence that the reasoning process mirrors that of a human chemist. Our results demonstrate that the latest reasoning models can, in some cases, perform advanced chemical reasoning.

\end{abstract}
\section{Introduction}
Large language models (LLMs) have emerged as powerful general-purpose tools across a variety of domains \cite{openai_gpt-4_2024}. Trained on extensive corpora of natural language, these models initially excelled at generating coherent text but often struggled with complex tasks that required deeper problem solving and reasoning. This limitation was at least partially addressed with the introduction of Chain-of-Thought (CoT) prompting, where the model is encouraged to conduct intermediate reasoning steps before arriving at a final answer \cite{wei2022chain,kojima2022large}. This has recently led to the development of ``reasoning models'' which are explicitly trained to optimize their CoT across a range of tasks \cite{deepseek-ai_deepseek-r1_2025}. By training LLMs with reinforcement learning, these models are able to develop reasoning strategies that are broadly applicable across multiple domains. 
Several reasoning models have been released recently, including OpenAI's o-series models \cite{noauthor_openai_nodate,openai_openai_2025}, DeepSeek R1 \cite{deepseek-ai_deepseek-r1_2025}, Google's Gemini Flash and Pro 2.5 \cite{noauthor_gemini_flash_2025, noauthor_gemini_pro_2025}, and Anthropic's Claude 3.7 Sonnet \cite{noauthor_claude_nodate}. These reasoning models have achieved substantial improvements on various benchmarks in mathematics, science, and coding \cite{noauthor_openai_nodate,noauthor_gemini_pro_2025,deepseek-ai_deepseek-r1_2025}.

The success of LLMs has prompted investigation into whether these models can aid scientific discovery by performing tasks such as generating novel hypotheses, planning experiments, and interpreting data \cite{boiko_autonomous_2023,gottweis_towards_2025}. LLMs are a potential collaborative tool for chemists and, if their performance continues to improve, may themselves lead to the discovery of novel chemical matter such as new materials, catalysts, and drugs \cite{white2023future}. There has been some initial exploration of the use of LLMs in chemistry \cite{caldasramos_review_2025}, with previous studies confirming LLMs have a broad understanding of chemistry \cite{mirza_are_2024,rein_gpqa_2023,phan_humanitys_2025,guo_what_2023,white_assessment_2023}. 

However, it has repeatedly been observed that LLMs struggle to interpret molecular structures, thus restricting their wider use in real-world scenarios \cite{jang_chain--thoughts_2024,m_bran_augmenting_2024,caldasramos_review_2025}. Researchers have primarily attempted to address this limitation through the development of ``agentic systems'' where LLMs are given access to external chemistry software to solve tasks that they can not directly perform \cite{m_bran_augmenting_2024,campbell_mdcrow_2025,boiko_autonomous_2023,caldasramos_review_2025}. 
While these systems are able to plan and execute multi-step workflows based on user-specified criteria, the developed workflows are typically not designed according to specific molecular structures of interest.
LLMs that can directly interpret molecular structures would substantially advance the capabilities of both standalone LLMs and LLM-based systems.

In this work, we investigated the ability of LLMs to autonomously perform a range of chemistry tasks that require molecular understanding, without assistance from external tools. To do so, we have constructed a new corpus of tasks, called ``ChemIQ'', specifically to test LLMs' understanding of organic molecules. Previous LLM benchmarks for chemistry are summarized in Appendix \ref{section:previous_benchmarks}; ChemIQ is distinct from these in three key aspects:
\begin{enumerate}[leftmargin=3em]
    \item The tasks included in our benchmark are highly focused on molecular comprehension, as opposed to previous benchmarks which combine questions from numerous chemistry disciplines. The majority of tasks in ChemIQ have not been considered in any other LLM evaluation. 
    \item All questions are algorithmically generated, allowing new questions to be readily produced. This allows failure modes of LLMs to be probed and benchmarks to be updated, for example by adding more complex questions as LLM capabilities increase, or periodically with new questions to ensure performance is not inflated due to data leakage. 
    \item ChemIQ more closely reflects potential use cases of LLMs than previous benchmarks. Most previous benchmarks contain exclusively \cite{rein_gpqa_2023,hendrycks_measuring_2021} or an overwhelming majority \cite{mirza_are_2024} of multiple choice questions. While these are valid for testing general knowledge, they can often be solved by elimination rather than direct reasoning. ChemIQ consists solely of short-answer questions, requiring the model to construct a solution as opposed to selecting from multiple options. As such, it more accurately reflects real-world tasks and should therefore be a more useful measure of performance \cite{alaa_medical_2025}.
\end{enumerate}

Using our ChemIQ benchmark, we find that the latest reasoning models now possess the ability to understand molecular structures and directly perform tasks requiring advanced chemical reasoning. Across 816 questions, the state-of-the-art reasoning models, o3-mini, Gemini Pro 2.5, and DeepSeek R1, answered 50-57\% of questions correctly in their highest reasoning level, with the overall accuracy depending on the level of reasoning used. This stands in stark contrast to the non-reasoning models, ChatGPT 4o, Gemini Flash 2.5 (in non-thinking mode), and DeepSeek V3, which answered only 3-7\% of questions correctly. 
For each task, we found similarities between the model’s reasoning process and that of a human chemist, suggesting a level of conceptual understanding of chemistry. 
Our results demonstrate that the latest reasoning models now have advanced chemical reasoning capabilities and are able to perform tasks that previously relied on expert human judgment. 

\begin{figure}[th]
  \centering
  \includegraphics[width=\linewidth, trim=20 550 20 0, clip]{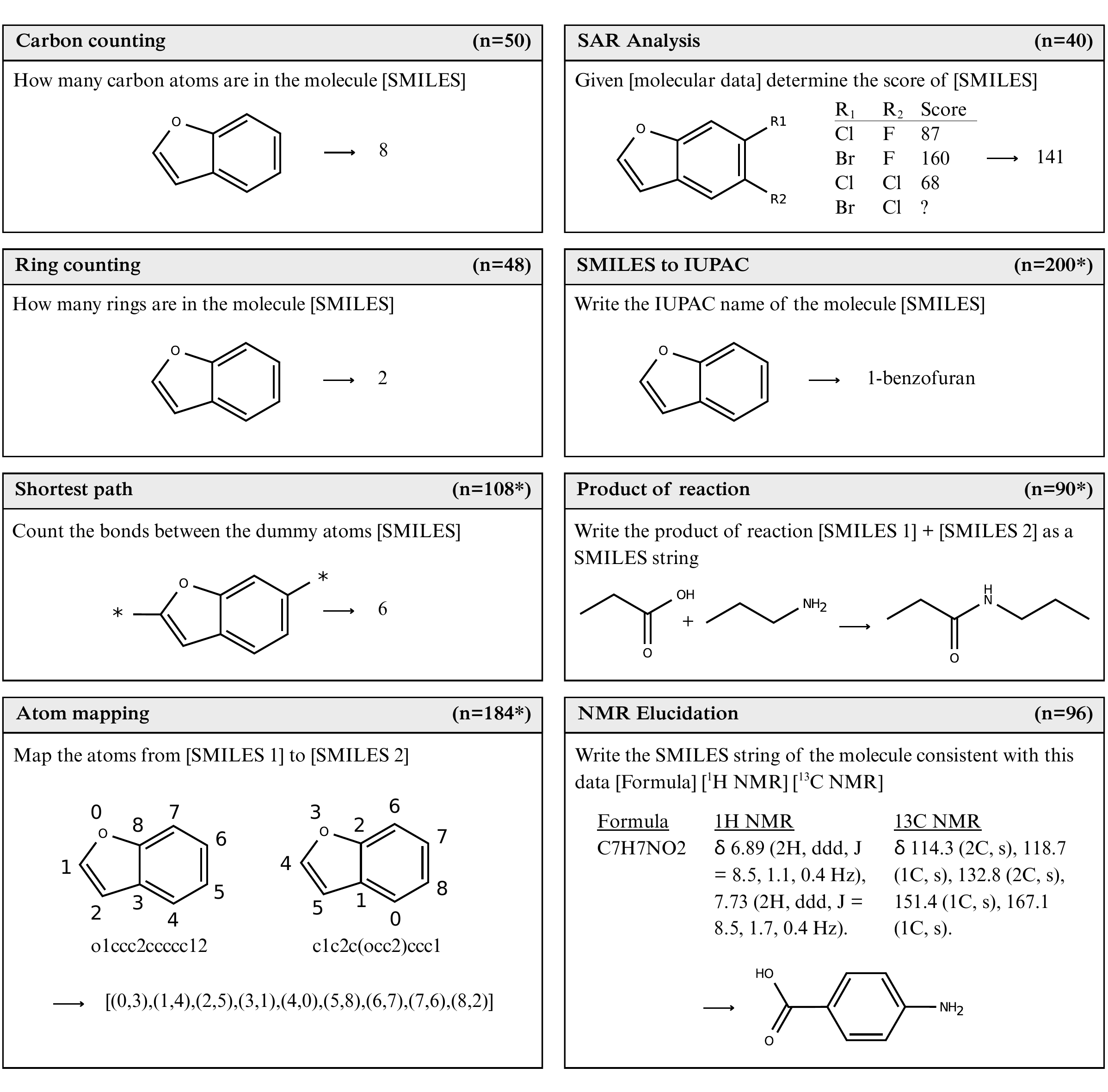}
  \caption{Question categories in the ChemIQ benchmark. The number of questions in each category is shown in the panel header, and * indicates that the set contains 50\% canonical and 50\% randomized SMILES. Full prompts and example reasoning excerpts can be found in Appendix \ref{app:reasoning_excerpts}.}
  \label{fig:all_tasks_visualised}
\end{figure}

\section{Construction of the ChemIQ benchmark}

ChemIQ is a novel benchmark developed to assess the chemical intelligence and understanding of LLMs. We designed the questions in ChemIQ to test the understanding of molecular structures and application of chemical reasoning as opposed to assessing general chemical knowledge or superficial pattern recognition.
ChemIQ focuses on three broad competencies: (1) interpreting molecular structures, (2) translating molecular structures to chemical concepts, and (3) reasoning about molecules using chemical theory. Competencies (1) and (2) represent foundational skills necessary for molecular comprehension, while competency (3) assesses more advanced chemical reasoning capabilities. 
We algorithmically generated a total of 816 questions across eight distinct categories, which are summarized in Figure \ref{fig:all_tasks_visualised}. 
Unless otherwise stated, we used molecules from the ZINC dataset \cite{sterling_zinc_2015} as examples of drug-like molecules.
We represented molecules using Simplified Molecular Input Line Entry System (SMILES) \cite{weininger_smiles_1988} as this format is widely used in cheminformatics and is a straightforward way for a human chemist to interact with LLM systems. Examples of full prompts and corresponding reasoning excerpts generated by o3-mini are provided in Appendix \ref{app:reasoning_excerpts}, and an example reasoning excerpt from Gemini Pro 2.5 is provided in Table \ref{tab:NMR_ZINC_reasoning}.

\subsection{Interpreting molecular structures}
To demonstrate even a basic understanding of molecules, the model must be able to count. For example, counting the number of atoms of each element is needed to determine the molecular formula. LLMs have previously been known to struggle with counting characters in text \cite{fu_why_2024}, suggesting potential challenges for interpreting SMILES strings. To test this ability, we asked the LLMs to count the number of carbon atoms and the number of rings in a molecule, properties that can be easily inferred from SMILES notation. 

A core skill for molecular interpretation is the ability to extract graph-based features from SMILES strings. Certain chemical features can be identified through simple pattern recognition (e.g. \texttt{C(=O)NC} indicates an amide group). Consequently, questions asking for the identification of functional groups within a molecule can give a misleading impression of molecular comprehension. To test the general ability to extract graph-based features from SMILES strings, we created a task that requires the LLM to determine the number of bonds on the shortest path between two randomly selected positions in a molecule. This requires a greater understanding of molecular structure than subgroup identification tasks.

Finally, we also constructed an ``atom mapping'' task to test global understanding of molecular structures. In this question, we assess whether the model can form an internal representation of molecules and recognize graph isomorphism. While a canonical SMILES format exists, a molecule can be represented by multiple SMILES strings. Furthermore, each atom can be assigned a unique index based on its position within a SMILES string. The LLM was given two randomized SMILES strings of the same molecule and was asked to provide the mapping of atoms from one representation to the other. Successfully answering this question requires an understanding of SMILES strings and molecular graphs, signaling the potential to perform more advanced reasoning on these structures.

\subsection{Translating molecular structures to chemical concepts}
A key objective for LLMs is to bridge the gap between computational representations of molecules and higher-level chemical terminology. Successfully doing so is necessary for any reasoning task that requires applying chemistry theory to molecular graphs, i.e. describing electronic effects, designing molecules, and performing structure elucidation. 

The International Union of Pure and Applied Chemistry (IUPAC) provides a standardized naming convention where molecules are described using common chemistry language; the ability to write the IUPAC names of molecules requires an understanding of molecular structures and their chemical features. Converting SMILES strings to IUPAC names has proven extremely challenging for LLMs, with most models achieving near-zero accuracy \cite{guo_what_2023,castro_nascimento_large_2023,yu_llasmol_2024,white_assessment_2023}. 

Previous benchmarks have typically assessed accuracy by comparing generated names to the exact name contained in PubChem \cite{kim_pubchem_2025}, which itself uses the OpenEye tool ``Lexichem'' \cite{noauthor_convert_nodate} to determine IUPAC names. While this approach is valid, we believe it is overly strict. Importantly, a single molecule can be identified by multiple valid IUPAC names, not only the one considered as ``standard''. While these non-standard names ignore conventions related to group priority, they nonetheless reflect an equivalent understanding of molecular structure. Therefore, we propose a modification of the SMILES to IUPAC task where a name is considered correct if it can be parsed to the intended structure using the Open Parser for Systematic IUPAC nomenclature (OPSIN) tool \cite{lowe_chemical_2011}.

\subsection{Chemical reasoning}
For an LLM to be used for novel chemical discovery, it must be capable of interpreting chemical trends in data. We assessed the potential to conduct such analysis by algorithmically constructing a series of structure-activity relationship Free-Wilson analysis questions. In each case, the LLM is provided data for seven molecules with different combinations of R-groups at three positions of a shared scaffold and is tasked with predicting the value for an unseen molecule. Further details of the scaffold and R-groups can be found in Appendix \ref{app:methods}.
This requires the LLM to be able to understand the differences between molecules, attribute values to these differences, and apply this knowledge to unseen examples.

We further tested chemical proficiency using a series of reaction prediction questions. These questions were constructed from nine common reaction classes, such as the copper-catalyzed click reaction and the Simmons-Smith cyclopropanation reaction. In each case, the LLM was prompted to write the product as a SMILES string. While the chosen templates reflect straightforward and unambiguous chemical reactions, correctly answering the reaction prediction questions requires substantial levels of chemical understanding and reasoning. 

Last, we assessed the ability of LLMs to interpret multiple sources of chemical evidence through the task of nuclear magnetic resonance (NMR) structure elucidation. While several machine learning models can provide helpful insights for this task, no current method can reliably interpret NMR spectra without expert validation \cite{hu_accurate_2024}. 
We provided the LLM with simulated NMR data, along with a molecular formula, and then prompted the model to generate the corresponding SMILES string. We assessed performance on molecules of up to ten heavy atoms from Huang \textit{et al.} using only 1D $^{1}$H and $^{13}$C NMR \cite{huang_framework_2021}. 
We additionally assessed the models using larger molecules of up to 30 heavy atoms from ZINC, providing both 1D $^{1}$H and $^{13}$C NMR, as well as 2D COSY, HSQC, and HMBC NMR spectra. 
This task requires pattern recognition in NMR data, mapping these patterns to molecular features, and subsequently constructing a coherent and complete molecular structure consistent with all sources of information.

\begin{figure}[!t]
  \centering
  \includegraphics[width=0.95\linewidth]{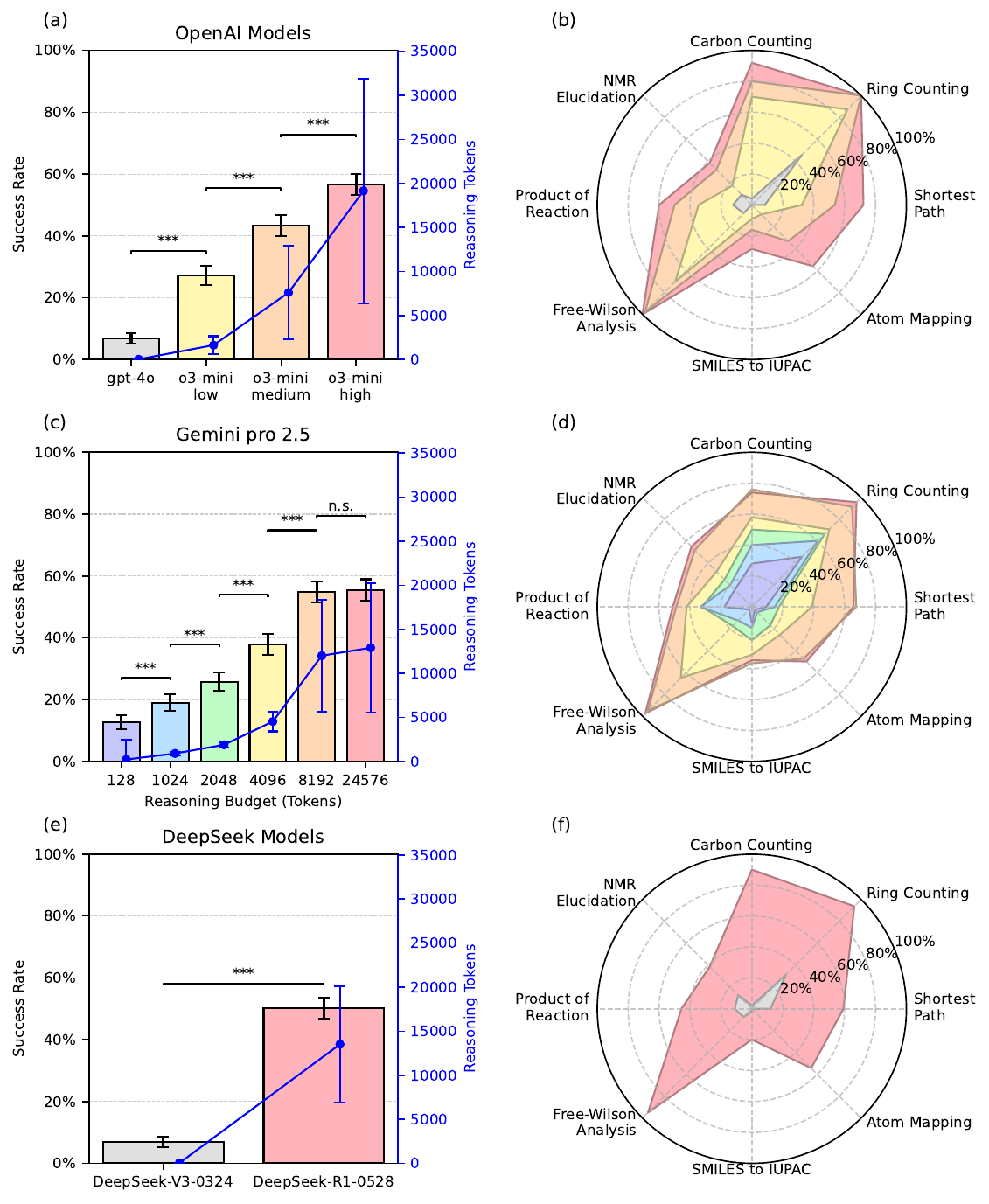}
  \caption{Summary of model performance. Results show a trend that increased reasoning correlates with higher success rates. A summary of question prompts is provided in Figure \ref{fig:all_tasks_visualised}, and full prompts with reasoning excerpts are listed in Appendix \ref{app:methods}. (a,c,e) Average success rates across all benchmark questions and the average number of reasoning tokens used per question. Vertical error bars show 95\% confidence interval. Adjacent bars were compared using a one-tailed McNemar test. Significance levels are denoted as: $^{*}p < 0.05$, $^{**}p < 0.01$, and $^{***}p < 0.001$. (b,d,f) Radar plots showing performance by question category. (a,b) OpenAI models GPT-4o (non-reasoning) and o3-mini (reasoning) (c,d) Google Gemini Pro 2.5 with varying reasoning budgets (e,f) DeepSeek models DeepSeek-V3-0324 (non-reasoning) and DeepSeek-R1-0528 (reasoning).}
  \label{fig:all_question_summary_results}
\end{figure}

\section{Results and Discussion}
\subsection{Experimental setup}
We evaluated the performance of multiple LLMs on our benchmark set of questions, ChemIQ, to assess their ability to understand molecular structures and perform chemical reasoning.
A key aim of our experiments was to assess the impact of ``reasoning'' on these capabilities. 
For many models, the extent of reasoning can be controlled using a ``reasoning budget'' parameter. We therefore chose to run o3-mini (available in ``low'', ``medium'', and ``high'' reasoning modes) and Gemini Pro 2.5 (with reasoning budgets between 128 and 32,768 tokens). We also ran Gemini Flash 2.5 which showed similar trends to Gemini Pro 2.5, but with weaker overall performance; the results of this model are listed in Appendix \ref{tab:gemini-flash-results-table}, but are not discussed further. We additionally ran DeepSeek-V3-0324 (non-reasoning) and DeepSeek-R1-0528 (reasoning) as examples of open-source models, and GPT-4o as a non-reasoning baseline for OpenAI models. 
By comparing the same model across multiple reasoning budgets, we can isolate the improvement gained from additional reasoning. 

To help understand how the models arrived at their answers, we examined the ``reasoning excerpts'' generated by the models. Of the models tested, DeepSeek R1 is the only model that provides a raw CoT; the other models do not disclose the raw CoT and instead provide a filtered summary of their reasoning. We use these reasoning excerpts as an indication of the reasoning performed by each model, although we note that reasoning traces are imperfect and may not describe the true method used to generate the final answers \cite{chen_reasoning_nodate}. A summary of each task is provided in Figure \ref{fig:all_tasks_visualised}, full prompts along with reasoning excerpts from o3-mini can be found in Appendix \ref{app:reasoning_excerpts}, and an example 2D NMR elucidation excerpt from Gemini Pro 2.5 can be found in Table \ref{tab:NMR_ZINC_reasoning}.
All benchmark questions in ChemIQ, answer checking scripts, and LLM outputs, including reasoning traces and excerpts, are available at https://github.com/oxpig/ChemIQ.

\subsection{Reasoning unlocks chemical intelligence}
The reasoning models substantially outperformed the non-reasoning models across the 816 questions in our benchmark. The reasoning models correctly answered 50-57\% of questions when run in the highest reasoning mode, whereas the non-reasoning models scored 3-7\%. 
Moreover, increased levels of reasoning enhanced performance across all tasks (Figure \ref{fig:all_question_summary_results}). 
However, minimal performance improvement was observed for Gemini Pro 2.5 at the three highest reasoning budgets, suggesting the model was unable to effectively utilize reasoning tokens beyond a certain threshold (Table \ref{tab:gemini-pro-results-table}).
We observed several model limitations beyond quantitative benchmark performance, discussed in Section~\ref{sec:results_limitations}.

The reasoning models answered the majority of numerical tasks involving counting and calculation correctly and demonstrated an ability to interpret graph-based features of molecules. On the SMILES to IUPAC task, the reasoning models achieved 29-44\% accuracy on molecules sampled from ZINC; to our knowledge, these are the first general-purpose LLMs to demonstrate meaningful success on this task. Additionally, all reasoning models solved the 1D NMR spectra of molecules up to 10 heavy atoms with 73-94\% accuracy, compared with 20-30\% accuracy of the non-reasoning models. The Gemini Pro 2.5 model exhibited the best performance in 2D NMR elucidation, correctly solving the structure of ten of the 50 molecules sampled from ZINC, one of which contained 25 heavy atoms. 

\begin{figure}[th]
  \centering
  \includegraphics[width=1\linewidth]{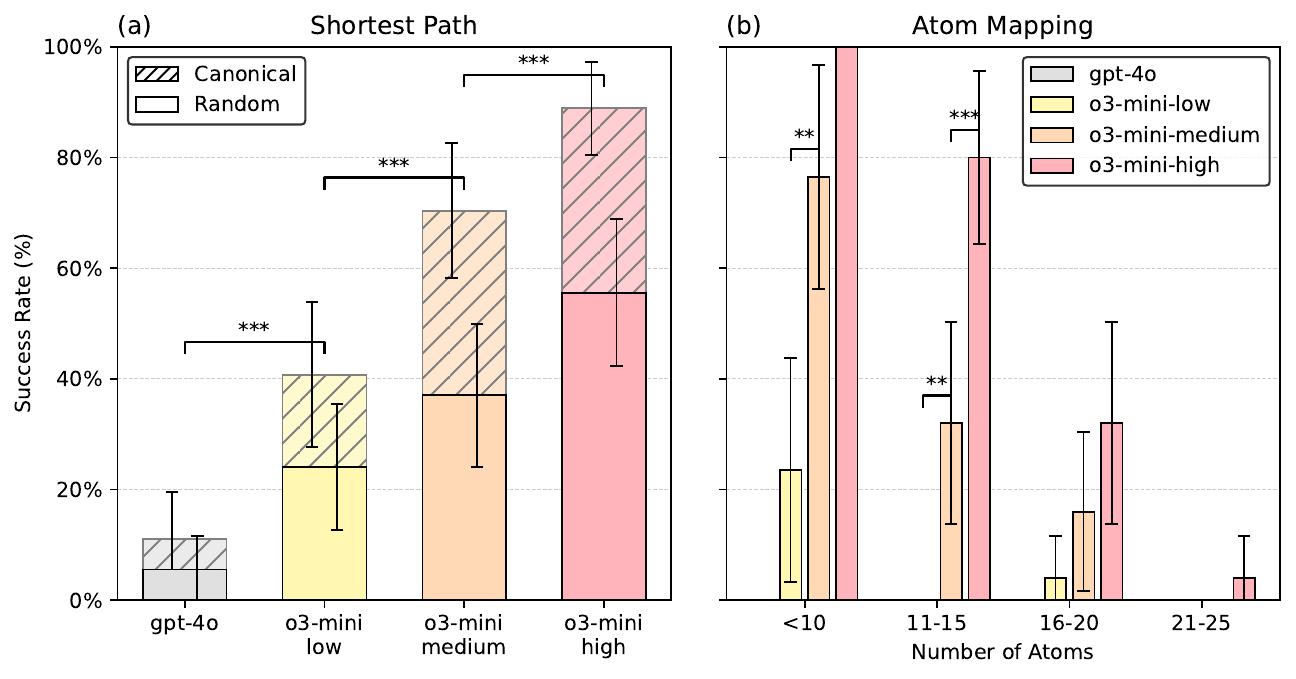}
  \caption{(A) Models were prompted to find the shortest distance between two atoms in a SMILES string. Performance was substantially lower when prompted with random SMILES as opposed to canonical SMILES. (B) Models were given two randomized SMILES strings representing the same molecules, and were prompted to map the atoms from one to the other. Performance decreased for larger molecules. Vertical error bars show 95\% confidence interval. Adjacent bars were compared using a one-tailed McNemar test. Significance levels are denoted as: $^{*}p < 0.05$, $^{**}p < 0.01$, and $^{***}p < 0.001$.}
  \label{fig:combined_shortest_path_atom_mapping}
\end{figure}

These results challenge the established view that LLMs cannot comprehend or write molecules in SMILES notation \cite{m_bran_augmenting_2024, guo_what_2023, bran_chemical_2025}. While the performance of the non-reasoning models is consistent with the conclusions from previous studies, the reasoning models can perform tasks that require comprehensive molecular and chemical understanding. 
However, across our entire benchmark, the best-performing model only successfully answered 57\% of questions. Thus, while the latest reasoning models show early signs of being able to perform chemistry tasks, they are not yet able to do so reliably. 

In the following sections, we focus primarily on the OpenAI models; however, similar trends were seen for all other models.

\subsection{Reasoning models understand molecular structures}
A fundamental requirement for understanding organic chemistry is the ability to interpret molecular structures. In this section, we discuss assessing molecular comprehension using a series of tasks that increase in complexity. As an initial test, we examined the ability of models to count characters in the SMILES string, which is a prerequisite for any other task involving the interpretation of SMILES notation. When tasked with carbon counting, GPT-4o only answered 4\% of questions correctly. The o3-mini model substantially improved on this, however was still imperfect, scoring between 70\% and 92\%. Performance on the ring counting task was higher, likely because the sampled molecules contained no more than six rings, which increased the chance of correct guesses and reduced overall task difficulty. These results are consistent with previous observations that LLMs struggle to count characters in text \cite{zhang2024counting}. This may limit the usefulness of SMILES as a chemical representation for current LLMs. 

We then examined the ability to understand molecular structures from SMILES strings using more complex questions beyond simple functional groups identification. The reasoning models demonstrated the ability to determine the shortest path between two atoms in a molecule (Figure \ref{fig:combined_shortest_path_atom_mapping}A), with o3-mini-high correctly answering 89\% of questions when molecules were written using canonical SMILES, and successfully reasoning about paths of up to 20 bonds. The non-reasoning GPT-4o model only scored 11\%. A challenging question is shown in Table \ref{tab:transcript_shortest_path} where the shortest path crossed a fused ring system. The reasoning excerpt suggests o3-mini-high constructed a graph representation and performed a graph search to solve this question, indicating the model can parse a SMILES string, build an internal graph representation, and then reason on it. 

Similarly, the reasoning models were able to answer ``atom mapping'' questions, where the same molecule was written as two different SMILES strings and the model was prompted to map the atoms from one to the other (Figure \ref{fig:combined_shortest_path_atom_mapping}B). The o3-mini-high model answered 52\% of questions correctly and successfully mapped molecules with up to 24 atoms, whereas GPT‑4o answered none of the questions correctly. As expected, the performance decreased as the number of atoms in the molecule increased, reflecting the greater complexity of atom mapping for larger molecules (Figure \ref{fig:combined_shortest_path_atom_mapping}B). However, the trend in success rate with increased reasoning suggests the ability to answer this question for larger molecules is not an inherent limitation of reasoning models, and may be improved further by using more reasoning tokens. 

An example reasoning excerpt for atom mapping is shown in Table \ref{tab:transcript_atom_mapping}. The excerpt suggests o3-mini-high used multiple layers of abstraction to solve this question, starting from a coarse-grained description of the molecule before focusing on individual atoms and specifically anchoring on distinctive heteroatoms to aid the search. This demonstrates that reasoning models have the capacity to interpret SMILES strings and navigate molecular graphs; however, this ability is not yet consistent or reliable. Furthermore, the low accuracy on structures larger than 20 heavy atoms implies that current models may face challenges in interpreting typical drug-sized molecules.

\subsection{Randomized SMILES are more challenging than canonical SMILES}
SMILES strings can be written in multiple forms depending on the atom ordering used to traverse the molecular graph \cite{arus-pous_randomized_2019}. If a model can reliably parse SMILES strings, the atom numbering should not impact the ability to interpret the molecular graph. We repeated the shortest path, atom mapping, SMILES to IUPAC, and reaction prediction questions using canonical and randomized SMILES, and found the reasoning models generally had lower performance when prompted with randomized SMILES (Table \ref{tab:ChatGPT-results-table}). 

This effect was most prominent for the shortest path question, where the performance of all o3-mini models dropped by approximately 40\% (Figure \ref{fig:combined_shortest_path_atom_mapping}A). This question, in particular, is made more challenging by considering randomized SMILES. Canonical SMILES strings are always written starting on a dummy atom, meaning the shortest path can often be inferred by the position of the second dummy atom in the string (together with consideration of branching and ring closures). In contrast, in randomized SMILES, the string-based distance between the dummy atoms typically differs significantly from the graph-based distance. 

For the atom mapping questions, we explored two approaches of generating multiple SMILES strings: fully randomized and semi-canonical, where SMILES were written by starting on a random atom but following canonical rules thereafter. When using semi-canonical SMILES, the performance of o3-mini-high improved to 69\%, compared to 52\% with fully randomized SMILES (Table \ref{tab:ChatGPT-results-table}). This may be due to entire substructures of the molecule being traversed in the same order, yielding pairs of SMILES strings with subsequences of identical characters, thus allowing mapping of these atoms directly from the SMILES string without considering the molecular graph. 

Overall, these results suggest that reasoning models may use heuristics to interpret SMILES strings, as opposed to parsing an entire molecular graph. Although the findings still demonstrate a basic understanding of SMILES notation, they show that reasoning models can not yet reliably parse SMILES strings and may struggle to interpret more complex molecular features that do not follow simple patterns. 

\begin{figure}[th]
  \centering
  \includegraphics[width=\linewidth, trim=0 0 0 0, clip]{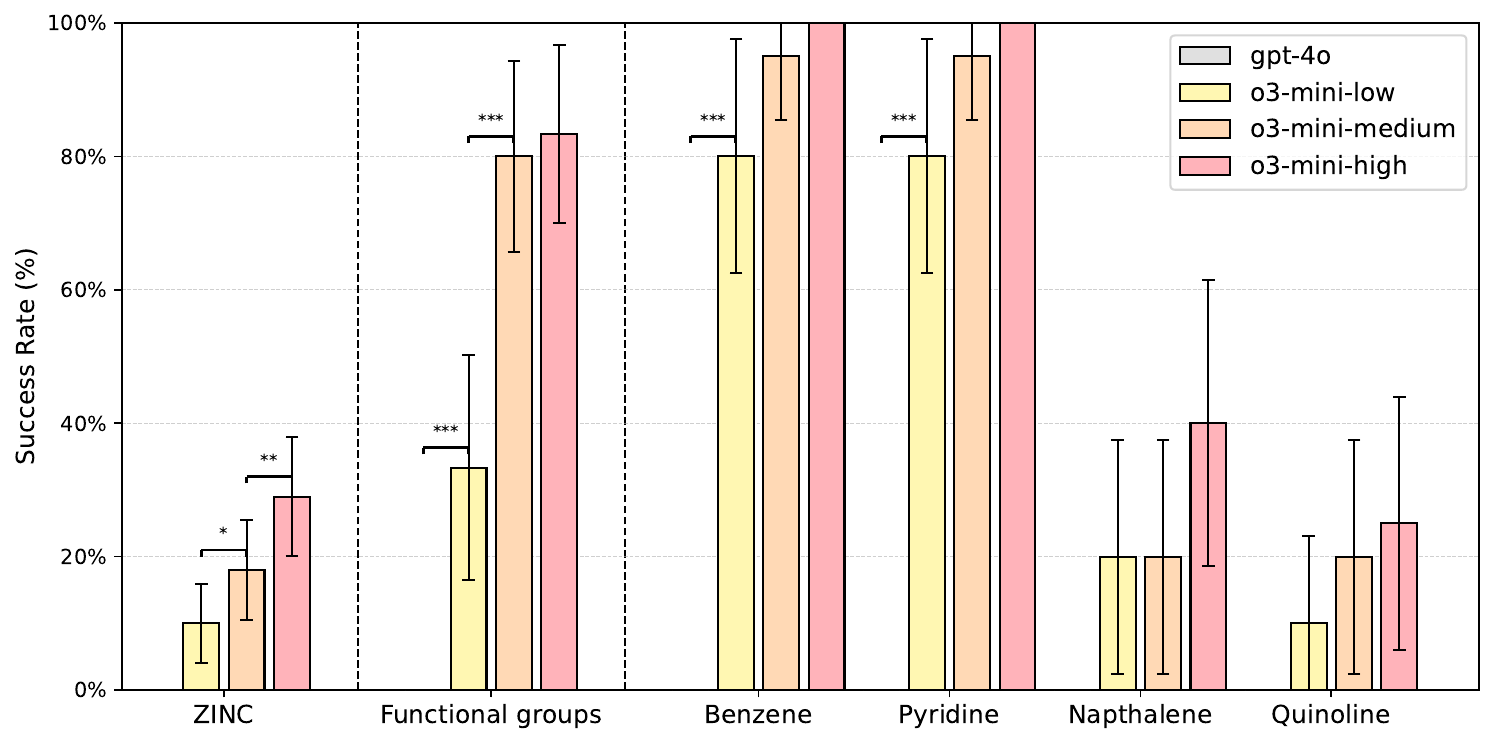}
  \caption{Performance on the SMILES to IUPAC questions. 
  The o3-mini model can successfully write IUPAC names of some molecules. The GPT-4o model answered all questions incorrectly. ZINC molecules were sampled from the ZINC dataset; correctly named molecules are shown in Figure \ref{fig:zinc_iupac_canonical_smiles}. Functional group questions were constructed by attaching 6 different functional groups to a benzene scaffold (Figure \ref{fig:smiles_to_iupac_functional_group_naming_mols}). Locant numbering questions were constructed by attaching 3 halogens to the listed scaffolds (Figure \ref{fig:smiles_to_iupac_locant_mols}). Names were accepted as correct if they could be parsed to the intended molecule by the OPSIN tool \cite{lowe_chemical_2011}. Results shown are for canonical SMILES; see Table \ref{tab:ChatGPT-results-table} and Table \ref{tab:additional_success_data} for randomized SMILES. Vertical error bars show 95\% confidence interval. Adjacent bars were compared using a one-tailed McNemar test. Significance levels are denoted as: $^{*}p < 0.05$, $^{**}p < 0.01$, and $^{***}p < 0.001$.}
  \label{fig:all_iupac_plt}
\end{figure}

\subsection{Reasoning models can write IUPAC names of molecules}
We then sought to test whether LLMs can map SMILES strings to chemical concepts. The most primitive requirement of this is to identify functional groups and describe their relationship to each other, which can be captured by the task of writing IUPAC names for molecules from SMILES strings. All previously tested models have achieved near-zero performance on IUPAC naming tasks \cite{guo_what_2023,castro_nascimento_large_2023,yu_llasmol_2024,white_assessment_2023}.

We found that reasoning-based LLMs are able to translate SMILES strings to IUPAC names for some molecules. 
The LLMs were initially tested using a sample of 100 molecules from the ZINC dataset. On these questions, GPT‑4o failed to generate any valid solutions, whereas o3-mini-low scored 10\%, o3-mini-medium 18\%, and o3-mini-high 29\%. The best performing model was Gemini Pro 2.5 which scored 35-44\% in the highest reasoning modes. In contrast to some other questions, the performance of o3-mini was similar when using randomized SMILES. There was a substantial overlap of molecules solved from both canonical and randomized SMILES, with o3-mini-high solving 16 molecules from both representations. These molecules are visualized in Figure \ref{fig:zinc_iupac_canonical_smiles} and Figure \ref{fig:zinc_iupac_random_smiles}. 

Closer inspection suggested that the ability to name molecules depended on the substructures present, as opposed to simply the size of the molecule. To assess the main error modes on this question, we algorithmically generated two sets of molecules: one to test the ability to identify and name functional groups, and the other to test the ability to describe their positions within a molecule. 

Using a set of 40 common functional groups, we randomly selected six groups and attached them at different positions on a benzene scaffold (example molecules are shown in Figure \ref{fig:smiles_to_iupac_functional_group_naming_mols}, further details are provided in Appendix \ref{app:methods}). 
o3-mini-low successfully produced IUPAC names for around 33\% of these molecules, while the medium and high models both achieved $\sim$80\% accuracy, demonstrating the models can accurately interpret functional groups from SMILES strings. We then tested the ability of the models to assign locant numbers by using a set of halogen-substituted scaffolds (Figure \ref{fig:smiles_to_iupac_locant_mols}). o3-mini in all three reasoning modes could accurately number benzene and pyridine scaffolds with at least 80\% success. The model often failed to number the fused ring systems of naphthalene and quinoline, scoring only $\sim$20\% on these questions. This could be due to fused ring systems being more challenging to parse from SMILES and IUPAC locant numbering of fused systems being inherently more complex. 

\begin{figure}[th]
  \centering
  \includegraphics[width=0.8\linewidth]{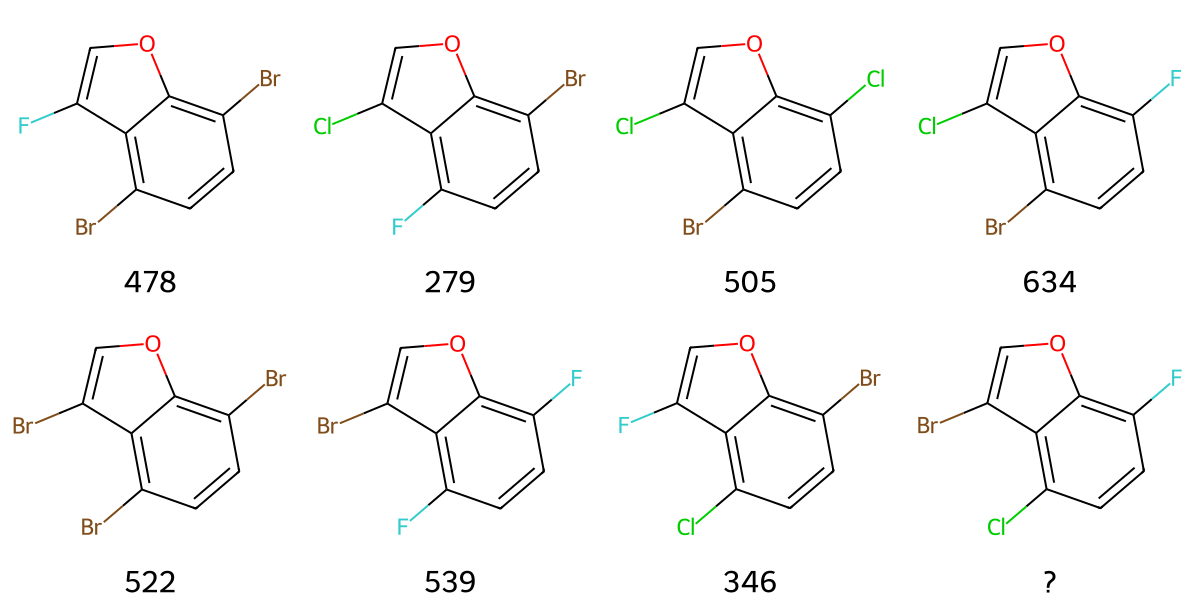}
  \caption{Example Free-Wilson analysis question. Molecules were constructed by attaching three halogens at three positions on a benzene scaffold. Each substituent at each position was assigned a random integer value, and the total score was calculated as the sum of these individual values. The task is to determine the score of the unknown molecule. The answer can be found in Table \ref{tab:transcript_SAR}.}
  \label{fig:sar_int}
\end{figure}

\subsection{Reasoning models have the potential to interpret chemical trends}
An important aspect of chemistry is the interpretation of chemical trends. We constructed a series of structure-activity relationship Free-Wilson analysis-type questions aimed at testing the ability to explain experimental measurements in terms of molecular features. An example question is shown in Figure \ref{fig:sar_int}. All reasoning models achieved 95-100\% on these questions. We also repeated these questions with the addition of noise to simulate more realistic data, and again, each reasoning model scored 95-100\%. One common strategy used to solve these questions was via a system of equations (Table \ref{tab:transcript_SAR}). These results suggest that the reasoning models are able to interpret mathematical relationships from molecular data and indicate that they may be able to analyze more complex experimental properties. 

\subsection{Reasoning models can write SMILES strings for simple reactions}
The ability to interpret and apply chemical concepts was evaluated through a series of reaction prediction tasks spanning nine reaction classes. For each reaction, the model was prompted to write the product as a SMILES string. To do so, the models must identify the reaction and apply the appropriate transformation. All models demonstrated some degree of chemical understanding, with the reasoning models achieving significantly higher performance. For example, across the nine reaction classes, GPT‑4o scored 18\%, o3-mini-low 36\%, o3-mini-medium 51\%, and o3-mini-high 56\%.

An example click reaction question is shown in Table \ref{tab:transcript_click_reaction}. The reasoning excerpt for this question appears to show o3-mini-high using (1) SMILES strings, (2) IUPAC names, (3) structural formulae, (4) atom numbers, and (5) chemical language interchangeably to arrive at the answer. Additionally, this excerpt highlights an instance where the model initially made a mistake by incorrectly writing the IUPAC name of a molecule, then shortly later performed an error check and corrected the issue. 

\subsection{Structure elucidation can be solved by reasoning models}
Finally, the most challenging question we tested the models with was NMR structure elucidation. Here, the models were provided with $^{1}$H and $^{13}$C NMR data, along with a molecular formula, and were asked to write the corresponding molecular structure as a SMILES string. We began by testing the LLMs on molecules containing up to 10 heavy atoms from \cite{huang_framework_2021}. All reasoning models solved the 1D NMR spectra of these molecules with 73-94\% accuracy, compared with 20-30\% accuracy of the non-reasoning models. The best performing model was Gemini Pro 2.5 which correctly determined the structures with 33-94\% accuracy depending on the reasoning budget. When tested on 2D NMR of molecules sampled from ZINC, Gemini Pro 2.5 successfully solved ten out of 50 structures, with one structure consisting of 25 heavy atoms (Table \ref{tab:NMR_ZINC_reasoning}).

The reasoning excerpts for these questions show a resemblance to that of a human chemist. An example small molecule question is shown in Table \ref{tab:transcript_nmr_small}. Here, o3-mini-high can be seen making a hypothesis that a carbon chemical shift corresponds to a carboxylic acid, before subsequently determining this to be inconsistent with the molecular formula. It then interprets methylene groups from the proton NMR and explains their deshielded chemical shift by proximity to an acetamide group. At this point, the model makes a comment that the integration of the proton peaks does not correspond to the number of protons given in the molecular formula, and proposes that the remaining protons must be exchangeable. Towards the end, the model proposes the IUPAC name of the structure, performs a double check, and then provides the correct SMILES string of the molecule. 

\begin{paddedtable}[th]
\caption{Excerpts of reasoning generated by Gemini Pro 2.5 on a 2D NMR structure elucidation question. The full prompt can be found in Appendix \ref{tab:full_zinc_prompt}. The model shows reasoning consistent with that of a human chemist. It combines information from the molecular formula and each NMR spectrum, considering factors such as chemical shifts, multiplicity, double bond equivalents, and cross peaks to deduce the structure of the molecule.}
\label{tab:NMR_ZINC_reasoning}
\begin{tabular}{|>{\columncolor{myLightGray}}m{0.1\textwidth}|m{0.9\textwidth}|}
\hline
Prompt & \begin{minipage}[t]{\linewidth}
{\small
Write the SMILES string of the molecule consistent with this data.\\

\textbf{Formula}: C19H19N3O3

\textbf{1H NMR}: $\delta$ 7.63 (ddd, J = 7.88, 1.41, 0.51 Hz, 1H), 7.6 (ddd, J = 8.14, 7.52, 1.41 Hz, 1H), [...]

\textbf{13C NMR}: $\delta$ 169.1 (1C, s), 161.4 (1C, s), 152.5 (1C, s), 140.5 (1C, s), 138.07 (1C, s), [...]

\textbf{COSY} ($\delta$H, $\delta$H): (7.6, 7.39), (7.6, 7.34), (7.63, 7.34), (7.07, 7.27), (7.48, 7.27), (3.85, 1.71) [...]

\textbf{HSQC} ($\delta$H, $\delta$C): (7.6, 135.15), (7.63, 129), (7.27, 128.9), (7.34, 126.75), (7.07, 123.58), [...]

\textbf{HMBC} ($\delta$H, $\delta$C): (4.66, 169.1), (7.63, 161.4), (3.85, 161.4), (3.85, 152.5), (4.66, 152.5), [...]\\

Only write the SMILES string. Do not write stereochemistry. Do not write any comments.
}
\end{minipage} \\
\hline
Reasoning & \begin{minipage}[t]{\linewidth}
{\small
[...] Looks like we've got a formula, C19H19N3O3, which means a lot of unsaturation – 12 degrees, to be exact. [...]\\

First, those aromatic signals in the 1H NMR, around 7-8 ppm, are screaming out for benzene rings. The integrations suggest one monosubstituted phenyl and one ortho-disubstituted one. Then there is a propyl chain (-CH2-CH2-CH3), which we can confirm with the COSY and HMBC. A singlet for a CH2 is a bit more tricky – deshielded, so it's probably close to something electronegative, or part of a pi system.\\

The 13C NMR gives us 19 carbons and a bunch of different signals, including ones for carbonyls, aromatic/vinylic carbons, and aliphatic carbons. With the HSQC, we can directly assign protons to carbons. The COSY confirms the propyl chain.\\

So, time to start piecing things together: we have the propyl chain and the phenyl group, which, are quite distinct. The HSQC gives correlations. We know the connections.\\

Let's work with fragments. We know we've got:
1. The Propyl chain.
2. The Phenyl group.
3. Ortho -disubstituted benzene ring.
4. Other atoms. Including a carbonyl group, and others with N and O.\\

Using the HMBC data, we can connect the fragments. The tricky part is working out what binds to what, and what core structure gives the overall formula. It also provides more clues. With H-4.66, it correlates to C-169.1, C-140.5, C-152.5, and this CH2 seems to be the key to the whole thing.\\

Given the N3 and O3, the high C values for the carbonyl groups point towards a heterocyclic core. We could try a pyrimidine-dione, or maybe even triazine-based structures. With a propyl chain attached. Maybe even a benzotriazole. HMBC correlations from the CH2 singlet, for instance, hint that it might be attached to a carbonyl group. Let's re-evaluate. C19H19N3O3. DBE=12.\\

Now, this is where it gets real tricky. The singlet for the CH2 really suggests a bridge, a connection point. A Phenylacetamide is one possibility, but it does not make sense in the ring. The Phenyl group attaches to the N, and then it will have a linker to the central parts, the CH2. The question is what is that connection.\\

After a lot of trial and error, looking at the HMBC data and trying to figure out all of the connections, I think that the structure is likely **1-((phenylcarbamoyl)methyl)-3-propylquinazoline-2,4-dione**.\\

[...] Now the SMILES: After many false starts, I finally have it. `c1ccc2c(c1)N(CC(=O)Nc2ccccc2)C(=O)N(CCC)C1O` [...]\\
}
\end{minipage} \\
\hline
Answer & {\small CCCN1C(=O)c2ccccc2N(CC(=O)Nc3ccccc3)C1=O }   \Large{\checkmark} \\
\hline
Visualized molecule & {\vspace{0.5em}\includegraphics[width=0.6\linewidth]{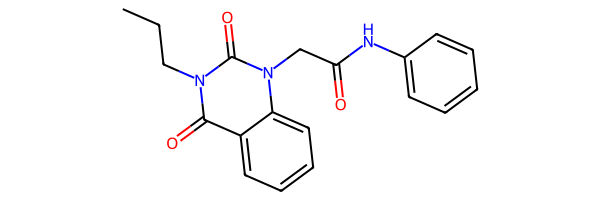}}\\
\hline
\end{tabular}
\end{paddedtable}

\FloatBarrier

The NMR elucidation excerpt from Gemini Pro 2.5 for the largest correctly solved ZINC molecule is given in Table \ref{tab:NMR_ZINC_reasoning} and shows advanced chemical reasoning. The model calculates the degree of unsaturation; identifies characteristic peaks; uses the HSQC to assign protons to specific carbon atoms; determines substructures using the COSY spectrum; combines substructures using the HMBC; focuses specifically on the HMBC of a CH$_{2}$ group to connect two separate spin systems of the molecule; writes the IUPAC name of the molecule; then generates the final SMILES string.  This result demonstrates a clear achievement in NMR elucidation and suggests that reasoning models can now address chemistry problems that previously relied on expert human judgment, marking a paradigm shift in computational chemistry.

\subsection{Model limitations}\label{sec:results_limitations}
Several limitations of the evaluated models beyond their quantitative benchmark performance were observed. 
The DeepSeek R1 model frequently failed to format its answer according to the given prompt. While most models occasionally suffered from this failure, this occurred significantly more frequently for DeepSeek R1. Implementing an algorithmic regular expression-based parser to extract the answer from the output significantly increased the performance of DeepSeek R1 from 24\% to 50\% accuracy, predominantly impacting the numerical and atom mapping questions (Table \ref{tab:deepseek-success-table}). Additionally, DeepSeek R1 exceeded its 32,000 token limit on 19 occasions, resulting in no generated response for these questions. 

The OpenAI o3-mini model frequently encountered unspecified API errors, which in particular affected the more challenging questions in the benchmark. For the 2D NMR questions, the initial API response rates were notably low: only 26/50 prompts succeeded for the o3-mini-high model and 35/50 for the o3-mini-medium model. After resubmitting the failed questions up to two more times, the o3-mini-high model still failed to give responses for nine 2D NMR questions. Given this error only occurred for the hardest questions in the benchmark, we speculate this error could be due to the underlying model filling the context window rather than a technical API error. 

The Gemini Pro 2.5 model regularly exceeded the allocated reasoning budget. For example, despite an 8,192-token allowance, the model used an average of approximately 12,000 reasoning tokens per response (Table \ref{tab:gemini-pro-token-table}). 

\section{Conclusion}
Our results show that language models that have been explicitly trained to reason now possess the capability to directly solve advanced chemistry problems without using external tools or requiring prompt engineering. Specifically, across eight distinct tasks ranging from atom counting to NMR elucidation, o3-mini, Gemini Pro 2.5, and DeepSeek R1 demonstrated the ability to interpret SMILES strings and reason on molecular graphs. For the first time, general-purpose LLMs were able to generate IUPAC names for molecules, with the reasoning models answering 29\% to 44\% of questions correctly when tested using the highest reasoning levels available. The performance of these reasoning models on the chemical reasoning tasks of NMR elucidation and reaction prediction signals a fundamental shift in the capabilities of LLMs. Our results strongly suggest that reasoning models now have the capacity to ``think'' about the structure of molecules, as opposed to just talking about them superficially. These findings represent a step change in the ability of language models to understand and reason about molecular structures.

Current reasoning models are still far from perfect. The models achieved only 74-92\% accuracy on the relatively simple task of carbon counting. All models were strongest at the Free-Wilson analysis (scoring 95-100\%), which predominantly requires mathematical reasoning. We expect continued advances in the general capabilities of LLMs to improve performance on domain-specific benchmarks such as ChemIQ. 
Furthermore, we expect that domain-specific models tailored for chemical reasoning, as well as LLM-based systems augmented with external tools, will deliver significant performance gains over general-purpose LLMs.

In our experiments, we observed improved performance for higher levels of reasoning; it is not clear to what extent this trend would continue if current models were allowed to reason for longer. We observed a plateau in performance for Gemini Pro 2.5 at the highest reasoning budgets, suggesting that there may be a limit to the performance improvement from allowing current reasoning models to use additional reasoning tokens. However, this coincided with the model choosing not to use more tokens, rather than using more reasoning tokens ineffectively, and there was no such plateau for o3-mini based on the three reasoning levels available. 

The same methodological advances enabling LLMs to solve complex chemistry problems risk lowering the barrier for producing chemical threats. The possibility that non-experts could leverage these capabilities to produce chemical threats is a serious concern \cite{openai_our_2025}, and one that has been largely overlooked in recent safety assessments that have prioritized biological risks \cite{openai_o3_2025,anthropic_claude_2025}. It is essential that researchers in this field are proactive in addressing the safety and ethical implications of this technology to prevent its misuse. 

The emerging capabilities of LLMs point to a new paradigm where LLMs serve as general-purpose chemical reasoning agents, combining formal theory and heuristic insights to solve a broad range of chemistry tasks. 
Such frameworks allow the integration of diverse information sources -- such as patents, primary literature, and domain knowledge -- into a single inference process, enabling context-aware decision-making and overcoming constraints that have historically limited conventional machine learning models. 
This could be transformational for data-limited chemical domains where developing task-specific models is currently often infeasible or insufficient.

\section*{Data and Code Availability}
All data (ChemIQ benchmark questions, answer checking scripts, and LLM outputs) for this study are available at https://github.com/oxpig/ChemIQ. 

\section*{Acknowledgments}
We thank Guy Durant and other members of the Oxford Protein Informatics Group for their helpful comments and discussions.

\bibliographystyle{sn-standardnature}
\bibliography{references}

\clearpage
\appendix

\section{Chemistry benchmarks for LLMs}
\label{section:previous_benchmarks}
Several benchmarks already exist to evaluate the performance of LLMs in chemistry. These benchmarks typically aggregate questions from multiple subfields and predominantly use a multiple choice format. While this approach can assess general chemical knowledge, it is less suitable for evaluating reasoning capabilities as questions can often be answered through elimination strategies rather than explicit chemical reasoning (see Table \ref{tab:IUPAC_to_SMILES_MCQ_reasoning} for an example from ChemBench). Furthermore, combining sub-fields within a single benchmark complicates the interpretation of specific capabilities of the LLM, and none of the existing benchmarks explicitly assess whether an LLM can interpret molecular structures. Below, we provide an overview of several prominent LLM benchmarks that assess topics in chemistry.

\textbf{ChemBench} \cite{mirza_are_2024} is a chemistry-specific benchmark consisting of 2634 questions, 2441 (c. 93\%) of which are multiple choice. These cover eight sub-categories: general chemistry ($n=149$), inorganic chemistry ($n=92$), materials science ($n=84$), organic chemistry ($n=429$), physical chemistry ($n=165$), technical chemistry ($n=40$), toxicity and safety ($n=675$), and medicinal chemistry preference ($n=1000$). To the best of our knowledge, the only question type in common with ChemIQ is ``SMILES to IUPAC''; however, ChemBench contains only 10 such questions, and all are multiple choice. Recent studies suggest that this benchmark has limited discriminatory power between reasoning and non-reasoning models. For example, the non-reasoning GPT‑4o model scored 61\%, whereas the reasoning o1 model scored 64\% \cite{mirza_are_2024}. Additionally, A separate study reported no difference in performance between o1, o3-mini-medium, and o3-mini-high on the ChemBench-mini subset \cite{wang_txgemma_nodate}. 

\textbf{Google-Proof Q\&A (GPQA)} \cite{rein_gpqa_2023} is a benchmark of 448 multiple choice questions, manually written by experts in biology, physics, and chemistry with the aim of being ``google-proof''. Each question has four options to select from, and is annotated with an estimate of difficulty level. All questions were assessed by multiple validators, and this was used to select a subset of 198 particularly challenging questions, known as ``GPQA diamond''. The benchmark contains 214 chemistry questions covering organic chemistry ($n=144$), general chemistry ($n=64$), inorganic chemistry ($n=3$), analytical chemistry ($n=2$), and physical chemistry ($n=1$).

\textbf{Humanity's last exam (HLE)} \cite{phan_humanitys_2025} is intended to be a challenging benchmark covering a broad range of subjects from the humanities and STEM. It contains 2,500 questions, with a mix of multi-modal questions (i.e. the prompt contains an image), multiple choice questions, and short-answer questions. Within the chemistry subset are 165 questions, 65 of which are multi-modal, and 43 are multiple choice. 

\textbf{Measuring Massive Multitask Language Understanding (MMLU)} \cite{hendrycks_measuring_2021} consists of 15,908 multiple choice questions from 57 different categories in humanities, social sciences, and STEM. These questions include 203 high school-level chemistry questions and 100 college-level chemistry questions. The complexity of these questions varies from knowledge recall to problem solving; however, the difficulty of individual questions is not quantified.  

\textbf{ChemLLMBench} \cite{guo_what_2023} was an early assessment of the potential of LLMs in chemistry. It consists of 800 questions, with 100 questions coming from each of eight subcategories: name prediction, property prediction, yield prediction, reaction prediction, reagent selection, retrosynthesis, text-based molecule design, and molecule captioning. These were assessed by a combination of binary classification, ranking, and short-answer questions. 

\clearpage
\section{Methods}\label{app:methods}
\subsection{Cheminformatics analysis}
All cheminformatics analysis was performed using the RDKit Python library (version 2022.09.5) \cite{noauthor_rdkit_nodate}.

\subsection{ZINC dataset}
The ZINC dataset used in this work refers to the ``molprop-250k-v2'' dataset accessed using the Polaris API \cite{sterling_zinc_2015,ash_practically_2024,wognum_call_2024}. This dataset was preprocessed by removing molecules with charged atoms, spiro centers, bridge head atoms, and more than 30 heavy atoms. The RingSystemLookup tool \cite{walters_patwalterspractical_cheminformatics_tutorials_2025} was used to remove molecules containing ring systems that appeared fewer than 100 times in the ChEMBL database, resulting in a final subset of $n = 141,562$ molecules. Stereochemistry was removed for each molecule, and the canonical SMILES string was used unless otherwise stated. 

\subsection{Token counting questions}
The number of carbon atoms in each molecule was calculated with \texttt{sum(1 for atom in mol.GetAtoms() if atom.GetSymbol() == 'C')}. Two ZINC molecules were sampled for each unique carbon atom count, resulting $n=50$ molecules ranging from three to 27 carbon atoms. The accepted answer for this question was the exact match of the integer. 

The number of rings in each molecule was calculated with \texttt{mol.GetRingInfo().NumRings()}. Eight ZINC molecules were sampled for each unique ring count, resulting in $n=48$ molecules ranging from one ring to six rings. The accepted answer for this question was the exact match of the integer. 

\subsection{Shortest path questions}
Each molecule in the ZINC subset was assigned a dummy atom at two random positions. The shortest path between these two points was calculated using the RDKit function \texttt{Chem.rdmolops.GetShortestPath(mol, dummy\_index\_0, dummy\_index\_1)}, subtracting one to determine the bond distance. Each molecule was written both as a canonical SMILES and a random SMILES string. 
Three molecules were sampled for every path length between three and 20 bonds, resulting in a total set of $n=54$ molecules. The accepted answer for this question was the exact match of the integer. 

\clearpage
\subsection{Atom mapping questions}
The atom mapping question only has a unique solution for molecules with no symmetric features, i.e. there is a single mapping of the molecule to itself. Molecules were identified as valid for this question if the substructure match of the maximum common substructure with itself resulted in a single match. By default, RDKit numbers atoms sequentially as they appear in the SMILES string; explicit hydrogens do not get numbered. The solution to this question was therefore determined by the MCS mapping of the two RDKit molecules. The pseudo-code snippet in Listing \ref{lst:atom_mapping} demonstrates this procedure. 

Using the subset of valid atom-mapping molecules, a maximum of five molecules were selected for each number of heavy atoms up to a maximum size of 25 heavy atoms. This resulted in $n=92$ molecules being selected. Each molecule was then used to create two separate questions, one using ``semi-canonical SMILES'' to represent the molecule and another using random SMILES. The ``semi-canonical SMILES'' follows a canonical walk around the molecule, however is rooted on a random atom. 
Thus, two semi-canonical SMILES strings are generally more similar than two random SMILES strings representing the same molecule.
The accepted answer in this question required all tuples to be correct, however they could be given in any order in the list. 

\begin{lstlisting}[caption={Pseudo code describing generation of atom mapping questions}, label={lst:atom_mapping}]
# Compute the MCS of the molecule with itself.
mcs_result = rdFMCS.FindMCS([mol, mol])
mcs_smarts = mcs_result.smartsString
mcs_mol = Chem.MolFromSmarts(mcs_smarts)

# Verify that the molecule is asymmetric.
matches = mol.GetSubstructMatches(mcs_mol, uniquify=False)
if len(matches) != 1:
    raise ValueError("The SMILES is not valid for this question due to symmetry (multiple mappings found).")

# random SMILES generated with doRandom=True and rootedAtAtom = -1.
# semi-canonical SMILES generated with doRandom=False and rootedAtAtom = random_index.
random_smiles1 = Chem.MolToSmiles(mol, rootedAtAtom=atom_idx1, doRandom=doRandom)
random_smiles2 = Chem.MolToSmiles(mol, rootedAtAtom=atom_idx2, doRandom=doRandom)

# Convert the generated SMILES back to molecules.
mol1 = Chem.MolFromSmiles(random_smiles1)
mol2 = Chem.MolFromSmiles(random_smiles2)

# Determine the atom mapping: for each atom in mol2, GetSubstructMatch returns the corresponding index in mol1.
mapping = mol1.GetSubstructMatch(mol2)

# Build a list of tuples: (atom index in smiles1, atom index in smiles2).
atom_mapping = [(mapping[i], i) for i in range(len(mapping))]
\end{lstlisting}

\clearpage
\subsection{SMILES to IUPAC}
For each question, the generated IUPAC name was assessed using the Open Parser for Systematic IUPAC nomenclature (OPSIN) API \cite{lowe_chemical_2011}. An IUPAC name was accepted as valid if it could be parsed to the intended structure. It is important to note that the generated IUPAC name only has to be self-consistent to be valid, as opposed to being the standardized/canonical IUPAC name; this is in contrast to previous benchmarks which have only accepted the standardized name as being correct. 

\paragraph{ZINC.} The ZINC dataset was split into $m=10$ bins by molecular weight, and 10 molecules were selected from each bin, resulting in a set of $n=100$ molecules.

\begin{figure}[H]
  \centering
  \includegraphics[width=1\linewidth]{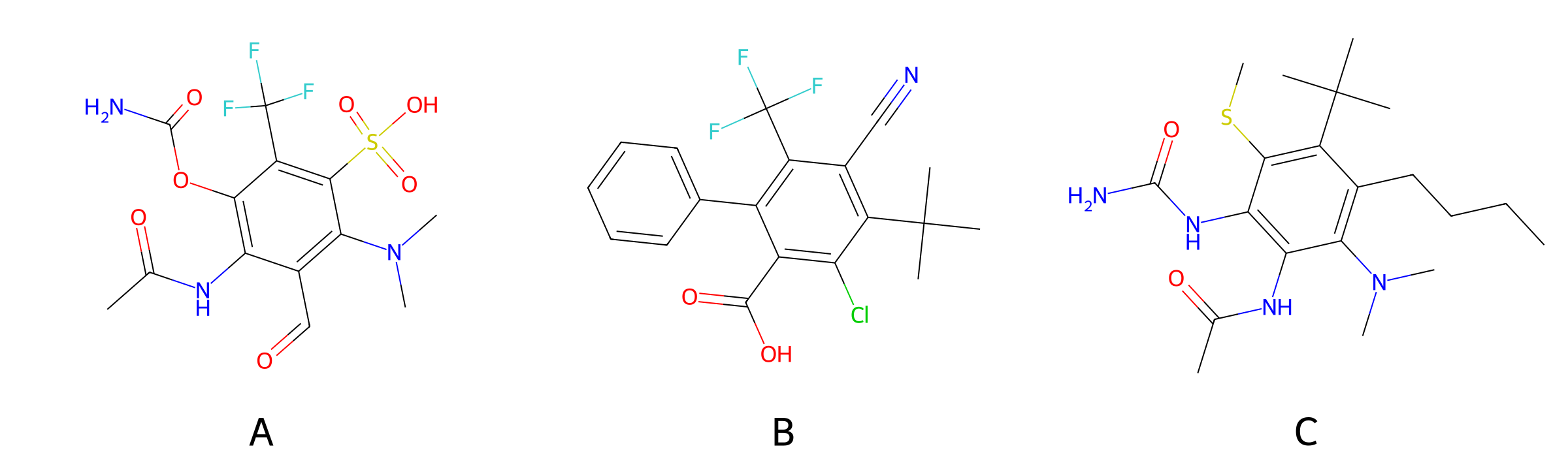}
  \vskip 0.5em
  {\footnotesize
    \begin{tabular}{@{}l@{\quad}l@{}}
      Molecule A: & \texttt{4‐acetamido‐3‐carbamoyloxy‐6‐(N,N‐dimethylamino)‐5‐formyl‐2‐(trifluoromethyl)}\\
      & \texttt{benzenesulfonic acid}\\
      Molecule B: & \texttt{2-Chloro-4-cyano-6-phenyl-3-(tert‐butyl)-5-(trifluoromethyl)benzoic acid}\\
      Molecule C: & \texttt{N-(2-acetamido-4-butyl-3-dimethylamino-6-(methylsulfanyl)-5-(tert-butyl)}\\
      & phenyl)urea\\
    \end{tabular}
  }
  \caption{Example functional group naming questions. Molecules generated by attaching 6 functional groups to a benzene scaffold. Listed IUPAC names were written by o3-mini-high and successfully parsed by OPSIN.}
  \label{fig:smiles_to_iupac_functional_group_naming_mols}
\end{figure}

\paragraph{Functional group.} A set of $n=40$ common functional groups were identified (listed in Table \ref{tab:functional_groups}). Functional groups containing heterocycle rings were excluded due to the known locant issues (tested separately). Molecules were generated by selecting six random functional groups from the set (without replacement) and attaching them to each of the six positions on benzene. $n=30$ molecules were generated by this method. Example molecules are shown in Figure \ref{fig:smiles_to_iupac_functional_group_naming_mols}.

\begin{figure}[H]
  \centering
  \includegraphics[width=1\linewidth]{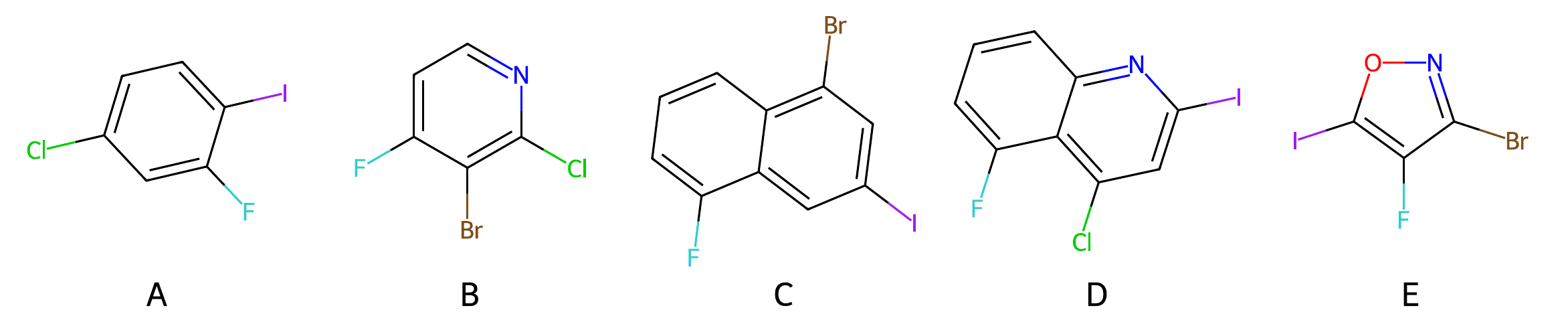}
  \vskip 0.5em
  {\footnotesize
    \begin{tabular}{@{}l@{\quad}l@{}}
      Molecule A: & \texttt{4-chloro-2-fluoro-1-iodobenzene}\\
      Molecule B: & \texttt{3-bromo-2-chloro-4-fluoropyridine}\\
      Molecule C: & \texttt{5‐bromo‐1‐fluoro‐7‐iodonaphthalene}\\
      Molecule D: & \texttt{2‐Iodo‐4‐chloro‐5‐fluoroquinoline}\\
      Molecule E: & \texttt{3-bromo-4-fluoro-5-iodoisoxazole}\\
    \end{tabular}
  }
  \caption{Example locant numbering questions. Scaffolds were selected as A) benzene B) pyridine C) naphthalene D) quinoline E) isoxazole. Three halogens were attached at 3 positions on the scaffold. Names were written by o3-mini-high and were successfully parsed by OPSIN.}
  \label{fig:smiles_to_iupac_locant_mols}
\end{figure}

\paragraph{Locant.} Benzene, pyridine, naphthalene, quinoline, and isoxazole were selected as scaffolds. Three halogens \{fluorine, chlorine, bromine, iodine\} were attached at three random positions on each scaffold. $n=20$ questions were generated for each scaffold. Example molecules shown in Figure \ref{fig:smiles_to_iupac_locant_mols}

\begin{table}[H]
  \centering
  \caption{Functional group used to generate functional group IUPAC naming molecules}
  \label{tab:functional_groups}
  \begin{tabular}{l|l|l}
   \toprule
    Common name        & Category      & SMILES           \\
    \midrule
    Fluoro             & Halogen       & [*]F             \\
    Chloro             & Halogen       & [*]Cl            \\
    Bromo              & Halogen       & [*]Br            \\
    Iodo               & Halogen       & [*]I             \\
    Trifluoromethyl    & Halogen       & [*]C(F)(F)F      \\
    Methyl             & Hydrocarbon   & [*]C             \\
    Ethyl              & Hydrocarbon   & [*]CC            \\
    n-Propyl           & Hydrocarbon   & [*]CCC           \\
    Isopropyl          & Hydrocarbon   & [*]C(C)C         \\
    n-Butyl            & Hydrocarbon   & [*]CCCC          \\
    Isobutyl           & Hydrocarbon   & [*]CC(C)C        \\
    sec-Butyl          & Hydrocarbon   & [*]C(C)CC        \\
    tert-Butyl         & Hydrocarbon   & [*]C(C)(C)C      \\
    Phenyl             & Hydrocarbon   & [*]c1ccccc1      \\
    Vinyl              & Hydrocarbon   & [*]C=C           \\
    Allyl              & Hydrocarbon   & [*]CC=C          \\
    Ethynyl            & Hydrocarbon   & [*]C\#C          \\
    Propargyl          & Hydrocarbon   & [*]CC\#C         \\
    Cyclopropyl        & Hydrocarbon   & [*]C1CC1         \\
    Cyclohexyl         & Hydrocarbon   & [*]C1CCCCC1      \\
    Amino              & Nitrogen      & [*]N             \\
    Secondary Amine    & Nitrogen      & [*]NC            \\
    Primary Amine      & Nitrogen      & [*]N(C)C         \\
    Nitrile            & Nitrogen      & [*]C\#N          \\
    Nitro              & Nitrogen      & [*]N(=O)=O       \\
    Hydroxyl           & Oxygen        & [*]O             \\
    Ether              & Oxygen        & [*]OC            \\
    Aldehyde           & Oxygen        & [*]C=O           \\
    Ketone             & Oxygen        & [*]C(=O)C        \\
    Carboxylic Acid    & Oxygen        & [*]C(=O)O        \\
    Ester              & Oxygen        & [*]C(=O)OC       \\
    Anhydride          & Oxygen        & [*]C(=O)OC(=O)C  \\
    Amide              & Oxygen        & [*]C(=O)NC       \\
    Reverse-Amide      & Oxygen        & [*]NC(=O)C       \\
    Carbamate          & Oxygen        & [*]OC(=O)N       \\
    Urea               & Oxygen        & [*]NC(=O)N       \\
    Thiol              & Sulfur        & [*]S             \\
    Thioether          & Sulfur        & [*]SC            \\
    Sulfonic Acid      & Sulfur        & [*]S(=O)(=O)O    \\
    Sulfonamide        & Sulfur        & [*]S(=O)(=O)N    \\
    \bottomrule
   \end{tabular}
\end{table}

\clearpage
\subsection{Structure activity relationship Free-Wilson analysis question generation}
The benzofuran core was chosen since each R group position can be unambiguously identified. Three random positions on the core were chosen as R group vectors. R groups were selected from the set of fluorine, bromine, and chlorine. Each R group at each vector was assigned a random integer value in the range (50,250). Molecules were generated by sampling an R group at each position, and the score of the molecule was set as the sum of these R groups. 

Questions were generated by constructing a system of equations and selecting a minimal set of R groups that yielded a full-rank matrix, guaranteeing that each variable (R group score) could be uniquely solved. The molecules selected in this step were provided in the prompt as context. An unknown molecule was randomly selected from the set of remaining molecules. 

A variation of this question was generated where noise was added to the total score of each molecule. Noise was sampled from a Gaussian distribution with mean zero, variance $2$. Considering the propagation of errors, the total variance in the predicted score for the unknown molecule is given by
\[
\operatorname{Var}(\hat{y}) = \sigma^2\left(1 + x^T\,(A^TA)^{-1}x\right),
\]
where \(\hat{y}\) is the predicted score of the unknown molecule and \(x\) is a vector indicating the presence or absence of each R group at the corresponding positions in the unknown molecule. The acceptable answer range was then defined by the 95\% confidence interval in order to account for noise in parameter estimation:
\[
\hat{y} \pm 1.96\sqrt{\sigma^2\left(1 + x^T\,(A^TA)^{-1}x\right)}.
\]

\subsection{Product of Reaction}
Nine undergraduate-level reactions were selected such that the product could be inferred from the reactants without information of the reaction conditions: Sn2, amide coupling, lithium grignard (1,2 addition at carbonyl), Michael addition (1,4 addition at $\alpha$,$\beta$-unsaturated carbonyl), Simmons-Smith (cyclopropanation), Wittig, Suzuki-Miyaura, Mizoroki-Heck, copper catalyzed click reaction. 

Each reaction was written as a reaction-template-like set of SMILES strings, with each reactant and product written with dummy atom attachment points indicating the position of R groups throughout the reaction. The set of R groups was selected such that they were unlikely to make the intended reaction in the question ambiguous. For each reaction template, R groups were randomly selected (without replacement) and attached to the template. Each reaction template was sampled five times with different R groups, yielding $n=45$ total synthetic reaction questions.
The model was provided with the reactants and prompted to write the product of reaction as a SMILES string.
Answers were assessed by reading the SMILES into RDKit, removing stereochemistry, canonicalizing the product, and then comparing this to the canonical SMILES of the expected product. Answers were accepted if both canonical SMILES strings matched. 

\subsection{NMR questions}
The set of molecules chosen for the ``small molecule'' NMR elucidation question were extracted from the supplementary information of Huang \textit{et al.} \cite{huang_framework_2021}. Their work used a neural network architecture to convert $^1$H and $^{13}$C NMR spectra to molecular structures. $n=23$ molecules were selected where the model did not rank the correct molecule as the top structure, and a second set of $n=23$ molecules were selected where the model was successful. This resulted in a set of $n=46$ molecules where the baseline Top-1 performance of the reference model was 50\%. All $^1$H and $^{13}$C NMR spectra were simulated using web tools from the nmrdb website www.nmrdb.org/ \cite{banfi_wwwnmrdborg_2008}. The outputted NMR text description (ie chemical shifts, peak integrals, and coupling constants) was used to construct the prompt. The LLM was prompted to generate the SMILES string of the molecule from this NMR data, along with the molecular formula. The answer was accepted as correct if the canonicalized SMILES string matched the expected answer. 

The ZINC 2D NMR molecules were selected by splitting the ZINC dataset into m=10 bins by number of heavy atoms, and selecting five molecules per bin, resulting in a set of $n=50$ molecules. The NMR spectra of each molecule were simulated using the nmrium NMR prediction web tool (www.nmrium.org/predict) \cite{Patiny2024}. The output of the web tool was downloaded as a ``.nmrium'' file which contained the data for 1D $^{1}$H and $^{13}$C NMR spectra, and the 2D COSY, HSQC, and HMBC NMR spectra. A python parser script was written to extract the NMR data. The diagonal and symmetric peaks in the COSY spectrum were removed, and the peaks of each spectrum were sorted in descending order by the chemical shift. Each spectrum was written into the prompt without further processing. The answer was accepted as correct if the canonicalized SMILES string matched the expected answer. 

\subsection{Statistical tests}
Models were compared to each other using the one-tailed McNemar test. The hypothesis was that the performance of models would increase with additional reasoning (ie GPT‑4o < o3-mini-low < o3-mini-medium < o3-mini-high). Only adjacent models in this series were tested. Horizontal staples on bar plots indicate the significance level for this test, with p values indicated with: $^{*}p < 0.05,\quad ^{**}p < 0.01,\quad ^{***}p < 0.001$. Adjacent vertical bars without a staple do not have a statistically significant difference in performance. 

The 95\% confidence interval of model success rate was calculated by the binomial proportion where \( \hat{p} \) is the success rate, and  \( n \) is the number of samples.
\[
\hat{p} \pm 1.96 \times \sqrt{\frac{\hat{p}(1-\hat{p})}{n}}
\]

The use of short answer questions generally means that a random model would be unable to guess the correct answer, or would achieve poor performance. For questions with integer solutions, the performance of a random model can be estimated by the ranges sampled to generate the questions. Considering the sampling ranges, the approximate baseline for each question is: carbon counting 1/25, ring counting 1/6, shortest path 1/18. The Free-Wilson analysis question does not have an obvious baseline, however would be close to 0\%. The SMILES to IUPAC, product of reaction, and NMR questions have a baseline of 0\%. The atom mapping question has a random baseline of 1/N! (where N is the number of atoms) meaning answers to this question are unlikely to be randomly guessed. 

\subsection{OpenAI ChatGPT models}
All questions were run using the OpenAI batch API. GPT‑4o was run using the 2024-11-20 snapshot. o3-mini was run using the 2025-01-31 snapshot, and by setting the reasoning effort parameter as ``low'', ``medium'', or ``high''. Prompts were submitted without ``function calling'' and without the ``code interpreter'' meaning the model could not use external tools to help answer questions. The raw chain-of-thought reasoning trace is not shared by OpenAI, however the ChatGPT web interface provides a summary of the reasoning process. The reasoning excerpts listed in this paper were constructed by manually prompting the web interface and copying the reasoning summaries. 

\subsection{Google Gemini models}
All Google Gemini models were run using the Google Vertex AI API. Questions were submitted using the batch API. The models used were ``gemini-2.5-flash'' (release date 2025-05-17, release stage GA) and ``gemini-2.5-pro'' (release date 2025-06-17, release stage GA). Reasoning budgets were set using the ``thinking\_budget'' parameter. Gemini Flash 2.5 is a ``hybrid'' model, meaning reasoning can be optionally disabled by setting the thinking budget to zero; this model was run with thinking budgets 0, 1024, 8192, and 24576. The Gemini Pro 2.5 model is a reasoning-only model; this model was run with thinking budgets 128, 1024, 2048, 4096, 8192, 24576, and 32768. In each case, the ``include\_thoughts'' parameter was set to True in order to obtain the reasoning summary. 

\subsection{DeepSeek models}
DeepSeek models were run using the DeepSeek API. The non-reasoning model ``DeepSeek-V3-0324'' was run using model alias ``deepseek-chat'', with default settings. The reasoning model ``DeepSeek-R1-0528'' was run with the alias ``deepseek-reasoner'' with the default context window of 32,000 tokens. 

\subsection{Answer parsing}
In some cases, the LLMs wrote a preamble such as ``my answer is...'' before writing the final answer. This predominantly affected the DeepSeek R1 model. Additionally, the DeepSeek model often wrote the atom mapping answer on separate lines, resulting in new line characters being present within the output. A parser script was written to extract the model's intended answer from the output text. This parser was used for all models; however, the parser primarily improved the performance of the DeepSeek R1 model.

\clearpage

\section{All Experimental Results}

\label{sec:additional-results}

\begin{table}[!th]
\centering
\caption{Success rates (\%) for OpenAI models on the ChemIQ benchmark. GPT-4o is a non-reasoning model; o3-mini is a reasoning model with low, medium, and high indicating the thinking budget for the model.}
\label{tab:ChatGPT-results-table}
\begin{tabular}{>{\raggedright\arraybackslash}p{4.5cm} c c c c}
\toprule
\textbf{Category} & gpt-4o & o3-mini-low & o3-mini-medium & o3-mini-high \\
\midrule
counting\_carbon (n=50) &   4.0\ &  70.0\ &  80.0\ &  92.0\ \\
\midrule
counting\_ring (n=48) &  45.8\ &  87.5\ & 100.0\ & 100.0\ \\
\midrule
shortest\_path\\\hspace{1em}canonical (n=54) &  11.1\ &  40.7\ &  70.4\ &  88.9\ \\
\hspace{1em}random (n=54) &   5.6\ &  24.1\ &  37.0\ &  55.6\ \\
\midrule
atom\_mapping\\\hspace{1em}semi-canonical (n=92) &   0.0\ &  12.0\ &  39.1\ &  62.0\ \\
\hspace{1em}random (n=92) &   0.0\ &   5.4\ &  27.2\ &  50.0\ \\
\midrule
smiles\_to\_iupac\\\hspace{1em}zinc\_canonical (n=100) &   0.0\ &  10.0\ &  18.0\ &  29.0\ \\
\hspace{1em}zinc\_random (n=100) &   0.0\ &   8.0\ &  14.0\ &  28.0\ \\
\midrule
sar\\\hspace{1em}integer (n=20) &   0.0\ &  75.0\ & 100.0\ & 100.0\ \\
\hspace{1em}noise (n=20) &  15.0\ &  65.0\ &  95.0\ & 100.0\ \\
\midrule
reaction\\\hspace{1em}synthetic\_canonical (n=45) &  17.8\ &  35.6\ &  51.1\ &  55.6\ \\
\hspace{1em}synthetic\_random (n=45) &   6.7\ &  33.3\ &  48.9\ &  64.4\ \\
\midrule
nmr\_elucidation\\\hspace{1em}small (n=46) &  19.6\ &  34.8\ &  65.2\ &  73.9\ \\
\hspace{1em}zinc\_2d (n=50) &   0.0\ &   2.0\ &   2.0\ &   6.0\ \\
\midrule
Total (n=816) &   6.9\ &  27.2\ &  43.4\ &  56.7\ \\
\bottomrule
\end{tabular}
\end{table}

\begin{table}[!th]
\centering
\caption{Success rates (\%) for Google Gemini Pro 2.5 on the ChemIQ benchmark. Gemini Pro 2.5 is a reasoning model; header is the thinking budget used for the model run.}
\label{tab:gemini-pro-results-table}
\begin{tabular}{>{\raggedright\arraybackslash}p{4cm} c c c c c c c}
\toprule
\textbf{Category} & 128 & 1024 & 2048 & 4096 & 8192 & 24576 & 32768 \\
\midrule
counting\_carbon (n=50) &  28.0\ &  40.0\ &  50.0\ &  58.0\ &  76.0\ &  74.0\ &  82.0\ \\
\midrule
counting\_ring (n=48) &  45.8\ &  60.4\ &  66.7\ &  70.8\ &  91.7\ &  95.8\ &  93.8\ \\
\midrule
shortest\_path\\\hspace{1em}canonical (n=54) &   9.3\ &  22.2\ &  27.8\ &  46.3\ &  79.6\ &  70.4\ &  72.2\ \\
\hspace{1em}random (n=54) &   9.3\ &   7.4\ &   7.4\ &  31.5\ &  55.6\ &  61.1\ &  53.7\ \\
\midrule
atom\_mapping\\\hspace{1em}semi-canonical (n=92) &   5.4\ &   6.5\ &  26.1\ &  33.7\ &  48.9\ &  53.3\ &  53.3\ \\
\hspace{1em}random (n=92) &   0.0\ &   3.3\ &   7.6\ &  17.4\ &  45.7\ &  46.7\ &  39.1\ \\
\midrule
smiles\_to\_iupac\\\hspace{1em}zinc\_canonical (n=100) &  16.0\ &  20.0\ &  25.0\ &  35.0\ &  42.0\ &  35.0\ &  44.0\ \\
\hspace{1em}zinc\_random (n=100) &   4.0\ &   7.0\ &  18.0\ &  27.0\ &  31.0\ &  34.0\ &  33.0\ \\
\midrule
sar\\\hspace{1em}integer (n=20) &   0.0\ &   0.0\ &  15.0\ &  65.0\ &  95.0\ & 100.0\ &  90.0\ \\
\hspace{1em}noise (n=20) &   5.0\ &  25.0\ &  20.0\ &  65.0\ &  95.0\ &  95.0\ & 100.0\ \\
\midrule
reaction\\\hspace{1em}synthetic\_canonical (n=45) &  28.9\ &  44.4\ &  40.0\ &  53.3\ &  55.6\ &  53.3\ &  62.2\ \\
\hspace{1em}synthetic\_random (n=45) &   6.7\ &  22.2\ &  24.4\ &  31.1\ &  42.2\ &  48.9\ &  48.9\ \\
\midrule
nmr\_elucidation\\\hspace{1em}small (n=46) &  32.6\ &  39.1\ &  50.0\ &  65.2\ &  89.1\ &  93.5\ &  84.8\ \\
\hspace{1em}zinc\_2d (n=50) &   0.0\ &   2.0\ &   2.0\ &   2.0\ &  18.0\ &  20.0\ &  20.0\ \\
\midrule
Total (n=816) &  12.6\ &  19.0\ &  25.7\ &  37.9\ &  54.8\ &  55.5\ &  55.5\ \\
\bottomrule
\end{tabular}
\end{table}

\begin{table}[!th]
\centering
\caption{Success rates (\%) for DeepSeek models on the ChemIQ benchmark. DeepSeek-V3-0324 is a non-reasoning model and DeepSeek-R1-0528 is a reasoning model. The DeepSeek R1 model often failed to write the answer in the format described in the prompt: a parser was written to extract the intended answer, resulting in improved performance. The ``without parsing'' is the performance of the model from the output only, and ``with parsing'' is the performance following algorithmic parsing of the answer from the output text.}
\label{tab:deepseek-success-table}
\begin{tabular}{>{\raggedright\arraybackslash}p{4.2cm} c c c}
\toprule
\textbf{Category} & DeepSeek-V3-0324 & \makecell{DeepSeek-R1-0528\\(without parsing)} & \makecell{DeepSeek-R1-0528\\(with parsing)}\\
\midrule
counting\_carbon (n=50) &   2.0\ &  68.0\ &  90.0\ \\
\midrule
counting\_ring (n=48) &  31.2\ &  37.5\ &  93.8\ \\
\midrule
shortest\_path\\\hspace{1em}canonical (n=54) &  13.0\ &   0.0\ &  66.7\ \\
\hspace{1em}random (n=54) &  11.1\ &   0.0\ &  51.9\ \\
\midrule
atom\_mapping\\\hspace{1em}semi-canonical (n=92) &   0.0\ &   5.4\ &  56.5\ \\
\hspace{1em}random (n=92) &   0.0\ &   4.3\ &  52.2\ \\
\midrule
smiles\_to\_iupac\\\hspace{1em}zinc\_canonical (n=100) &   2.0\ &  30.0\ &  30.0\ \\
\hspace{1em}zinc\_random (n=100) &   1.0\ &  10.0\ &  10.0\ \\
\midrule
sar\\\hspace{1em}integer (n=20) &   0.0\ &  20.0\ &  95.0\ \\
\hspace{1em}noise (n=20) &  15.0\ &  45.0\ &  95.0\ \\
\midrule
reaction\\\hspace{1em}synthetic\_canonical (n=45) &  15.6\ &  53.3\ &  53.3\ \\
\hspace{1em}synthetic\_random (n=45) &   6.7\ &  37.8\ &  37.8\ \\
\midrule
nmr\_elucidation\\\hspace{1em}small (n=46) &  26.1\ &  76.1\ &  76.1\ \\
\hspace{1em}zinc\_2d (n=50) &   0.0\ &   4.0\ &   4.0\ \\
\midrule
Total (n=816) &   7.0\ &  23.5\ &  50.2\ \\
\bottomrule
\end{tabular}
\end{table}

\begin{table}[!th]
\centering
\caption{Success rates (\%) for Google Gemini Flash 2.5 on the ChemIQ benchmark. Gemini Flash 2.5 is a hybrid model capable of both non-reasoning and reasoning modes. Header is the thinking budget used for the model run: a thinking budget of 0 means the model was run in the non-reasoning mode}
\label{tab:gemini-flash-results-table}
\begin{tabular}{>{\raggedright\arraybackslash}p{4.5cm} c c c c}
\toprule
\textbf{Category} & 0 & 1024 & 8192 & 24576 \\
\midrule
counting\_carbon (n=50) &   2.0\ &  42.0\ &  64.0\ &  58.0\ \\
\midrule
counting\_ring (n=48) &   8.3\ &  77.1\ &  93.8\ &  91.7\ \\
\midrule
shortest\_path\\\hspace{1em}canonical (n=54) &   1.9\ &  29.6\ &  44.4\ &  46.3\ \\
\hspace{1em}random (n=54) &   1.9\ &  27.8\ &  27.8\ &  37.0\ \\
\midrule
atom\_mapping\\\hspace{1em}semi-canonical (n=92) &   1.1\ &   5.4\ &  26.1\ &  27.2\ \\
\hspace{1em}random (n=92) &   0.0\ &   1.1\ &   8.7\ &  18.5\ \\
\midrule
smiles\_to\_iupac\\\hspace{1em}zinc\_canonical (n=100) &   0.0\ &   7.0\ &  11.0\ &  15.0\ \\
\hspace{1em}zinc\_random (n=100) &   1.0\ &   4.0\ &  14.0\ &  11.0\ \\
\midrule
sar\\\hspace{1em}integer (n=20) &   0.0\ &  85.0\ &  95.0\ &  90.0\ \\
\hspace{1em}noise (n=20) &  10.0\ &  70.0\ &  90.0\ &  85.0\ \\
\midrule
reaction\\\hspace{1em}synthetic\_canonical (n=45) &   6.7\ &  26.7\ &  26.7\ &  28.9\ \\
\hspace{1em}synthetic\_random (n=45) &   0.0\ &  24.4\ &  28.9\ &  35.6\ \\
\midrule
nmr\_elucidation\\\hspace{1em}small (n=46) &  30.4\ &  19.6\ &  58.7\ &  76.1\ \\
\hspace{1em}zinc\_2d (n=50) &   0.0\ &   2.0\ &   2.0\ &   2.0\ \\
\midrule
Total (n=816) &   3.4\ &  20.8\ &  32.2\ &  35.0\ \\
\bottomrule
\end{tabular}
\end{table}

\begin{table}[!th]
\centering
\caption{Total output tokens for OpenAI models on the ChemIQ benchmark. GPT-4o is a non-reasoning model; o3-mini is a reasoning model with low, medium, and high indicating the thinking budget for the model.}
\label{tab:ChatGPT-token-table}
\begin{tabular}{>{\raggedright\arraybackslash}p{4.5cm} c c c c}
\toprule
\textbf{Category} & gpt-4o & o3-mini-low & o3-mini-medium & o3-mini-high \\
\midrule
counting\_carbon (n=50) &   2.0 & 1582.4 & 4506.2 & 10209.9 \\
\midrule
counting\_ring (n=48) &   2.0 & 658.3 & 2725.1 & 8258.7 \\
\midrule
shortest\_path\\\hspace{1em}canonical (n=54) &   2.0 & 1748.4 & 4339.4 & 10348.5 \\
\hspace{1em}random (n=54) &   2.0 & 2171.6 & 6530.8 & 15697.8 \\
\midrule
atom\_mapping\\\hspace{1em}semi-canonical (n=92) &  83.2 & 1988.5 & 12344.5 & 27165.8 \\
\hspace{1em}random (n=92) &  85.5 & 1530.2 & 14690.3 & 35599.1 \\
\midrule
smiles\_to\_iupac\\\hspace{1em}zinc\_canonical (n=100) &  33.1 & 1690.3 & 6374.5 & 17962.7 \\
\hspace{1em}zinc\_random (n=100) &  35.0 & 1528.3 & 7358.9 & 21630.9 \\
\midrule
sar\\\hspace{1em}integer (n=20) &   2.0 & 1507.5 & 2896.3 & 4499.4 \\
\hspace{1em}noise (n=20) &   4.0 & 1468.7 & 4004.3 & 6331.6 \\
\midrule
reaction\\\hspace{1em}synthetic\_canonical (n=45) &  20.3 & 1051.2 & 4172.1 & 10745.8 \\
\hspace{1em}synthetic\_random (n=45) &  21.5 & 1271.0 & 4855.7 & 13950.9 \\
\midrule
nmr\_elucidation\\\hspace{1em}small (n=46) &  12.5 & 2113.2 & 6167.0 & 13706.9 \\
\hspace{1em}zinc\_2d (n=50) &  27.4 & 2102.6 & 11496.3 & 35292.1 \\
\midrule
Total (n=816) &  32.7 & 1635.2 & 7605.9 & 19147.8 \\
\bottomrule
\end{tabular}
\end{table}

\begin{table}[!th]
\centering
\caption{Total output tokens for Google Gemini Pro 2.5 on the ChemIQ benchmark. Gemini Pro 2.5 is a reasoning model; header is the thinking budget used for the model run.}
\label{tab:gemini-pro-token-table}
\begin{tabular}{>{\raggedright\arraybackslash}p{4.2cm} c c c c c c c}
\toprule
\textbf{Category} & 128 & 1024 & 2048 & 4096 & 8192 & 24576 & 32768 \\
\midrule
counting\_carbon (n=50) &  83.5 & 845.9 & 1893.1 & 4957.5 & 8912.8 & 9248.7 & 9409.6 \\
\midrule
counting\_ring (n=48) &  89.6 & 820.4 & 1831.8 & 4333.2 & 8801.5 & 8729.0 & 8724.7 \\
\midrule
shortest\_path\\\hspace{1em}canonical (n=54) &  87.4 & 846.5 & 1882.6 & 4892.8 & 9893.3 & 9994.7 & 10747.7 \\
\hspace{1em}random (n=54) &  92.3 & 817.4 & 1820.2 & 5205.6 & 13363.9 & 14256.8 & 14064.6 \\
\midrule
atom\_mapping\\\hspace{1em}semi-canonical (n=92) & 242.5 & 1060.8 & 2058.5 & 4332.8 & 12737.2 & 14598.1 & 14686.2 \\
\hspace{1em}random (n=92) & 236.4 & 1039.3 & 2051.4 & 4553.0 & 15749.4 & 17987.5 & 19034.5 \\
\midrule
smiles\_to\_iupac\\\hspace{1em}zinc\_canonical (n=100) & 116.1 & 852.5 & 1852.6 & 4519.9 & 10820.8 & 11270.0 & 11995.6 \\
\hspace{1em}zinc\_random (n=100) & 124.3 & 871.5 & 1864.6 & 4571.1 & 11875.5 & 12775.1 & 12694.6 \\
\midrule
sar\\\hspace{1em}integer (n=20) &  71.8 & 876.8 & 1886.3 & 4036.9 & 7262.7 & 8003.1 & 9911.9 \\
\hspace{1em}noise (n=20) & 102.6 & 927.5 & 1882.3 & 4541.2 & 9820.1 & 10416.0 & 11051.9 \\
\midrule
reaction\\\hspace{1em}synthetic\_canonical (n=45) & 101.3 & 837.0 & 1741.1 & 4147.4 & 9272.2 & 8573.3 & 9688.3 \\
\hspace{1em}synthetic\_random (n=45) & 111.8 & 882.4 & 1789.8 & 4279.2 & 8905.0 & 10261.2 & 9766.8 \\
\midrule
nmr\_elucidation\\\hspace{1em}small (n=46) & 147.0 & 845.8 & 1770.2 & 4505.5 & 9829.3 & 9816.1 & 10282.3 \\
\hspace{1em}zinc\_2d (n=50) & 1458.8 & 970.8 & 1809.9 & 4318.0 & 23461.0 & 25402.0 & 25421.0 \\
\midrule
Total (n=816) & 219.4 & 904.5 & 1885.5 & 4536.4 & 12008.9 & 12905.6 & 13283.4 \\
\bottomrule
\end{tabular}
\end{table}

\begin{table}[!th]
\centering
\caption{Total output tokens for DeepSeek models on the ChemIQ benchmark. DeepSeek-V3-0324 is a non-reasoning model and DeepSeek-R1-0528 is a reasoning model.}
\label{tab:deepseek-token-table}
\begin{tabular}{>{\raggedright\arraybackslash}p{4.2cm} c c}
\toprule
\textbf{Category} & DeepSeek-V3-0324 & DeepSeek-R1-0528 \\
\midrule
counting\_carbon (n=50) &   1.0 & 13449.8 \\
\midrule
counting\_ring (n=48) &   1.0 & 15331.1 \\
\midrule
shortest\_path\\\hspace{1em}canonical (n=54) &   1.0 & 15394.5 \\
\hspace{1em}random (n=54) &   1.0 & 16090.2 \\
\midrule
atom\_mapping\\\hspace{1em}semi-canonical (n=92) &  87.0 & 16200.4 \\
\hspace{1em}random (n=92) &  95.7 & 19263.8 \\
\midrule
smiles\_to\_iupac\\\hspace{1em}zinc\_canonical (n=100) &  33.2 & 9410.8 \\
\hspace{1em}zinc\_random (n=100) &  36.7 & 10284.5 \\
\midrule
sar\\\hspace{1em}integer (n=20) &   1.0 & 11191.1 \\
\hspace{1em}noise (n=20) &   3.0 & 11483.1 \\
\midrule
reaction\\\hspace{1em}synthetic\_canonical (n=45) &  18.1 & 7942.3 \\
\hspace{1em}synthetic\_random (n=45) &  19.5 & 9626.0 \\
\midrule
nmr\_elucidation\\\hspace{1em}small (n=46) &   9.2 & 9951.7 \\
\hspace{1em}zinc\_2d (n=50) &  23.2 & 19635.5 \\
\midrule
Total (n=816) &  33.5 & 13510.3 \\
\bottomrule
\end{tabular}
\end{table}

\begin{table}[ht]
\centering
\caption{Total output tokens for Google Gemini Flash 2.5 on the ChemIQ benchmark. Gemini Flash 2.5 is a hybrid model capable of both non-reasoning and reasoning modes. Header is the thinking budget used for the model run: a thinking budget of 0 means the model was run in the non-reasoning mode.}
\label{tab:gemini-flash-token-table}
\begin{tabular}{>{\raggedright\arraybackslash}p{4.5cm} c c c c}
\toprule
\textbf{Category} & 0 & 1024 & 8192 & 24576 \\
\midrule
counting\_carbon (n=50) &   1.9 & 1344.4 & 5695.7 & 6320.5 \\
\midrule
counting\_ring (n=48) &   1.0 & 1150.3 & 6164.2 & 7163.5 \\
\midrule
shortest\_path\\\hspace{1em}canonical (n=54) &   2.0 & 1915.1 & 8577.7 & 9563.7 \\
\hspace{1em}random (n=54) &   2.0 & 2787.9 & 9917.4 & 10760.4 \\
\midrule
atom\_mapping\\\hspace{1em}semi-canonical (n=92) & 108.9 & 1066.7 & 12676.8 & 16311.6 \\
\hspace{1em}random (n=92) & 116.3 & 949.6 & 15251.7 & 18975.9 \\
\midrule
smiles\_to\_iupac\\\hspace{1em}zinc\_canonical (n=100) & 698.7 & 877.5 & 6075.9 & 6508.8 \\
\hspace{1em}zinc\_random (n=100) &  43.5 & 840.5 & 6992.9 & 7712.9 \\
\midrule
sar\\\hspace{1em}integer (n=20) &   3.0 & 4815.1 & 9362.4 & 10203.8 \\
\hspace{1em}noise (n=20) &   6.0 & 3646.6 & 10075.8 & 10184.4 \\
\midrule
reaction\\\hspace{1em}synthetic\_canonical (n=45) &  18.4 & 1014.5 & 5592.2 & 5788.5 \\
\hspace{1em}synthetic\_random (n=45) &  20.2 & 1781.2 & 6823.3 & 6197.9 \\
\midrule
nmr\_elucidation\\\hspace{1em}small (n=46) &  11.3 & 776.5 & 8836.0 & 10047.9 \\
\hspace{1em}zinc\_2d (n=50) &  26.8 & 843.4 & 25449.1 & 31794.6 \\
\midrule
Total (n=816) & 121.4 & 1356.2 & 9904.5 & 11550.3 \\
\bottomrule
\end{tabular}
\end{table}


\clearpage

\begin{table}[ht]
\centering
\caption{Success rate (\%) by model for additional questions. Stated ± is 95\% confidence interval (binomial proportion).}
\label{tab:additional_success_data}
\resizebox{\linewidth}{!}{
\begin{tabular}{llllll}
\toprule
                          Question Category &      GPT‑4o & GPT‑4o-tags & o3-mini-low & o3-mini-medium & o3-mini-high \\
\midrule
      SMILES to IUPAC\\ \hspace{2em}Functional groups$^c$ ($n=30$) &   0.0 ± 0.0 &   0.0 ± 0.0 & 33.3 ± 16.9 &    80.0 ± 14.3 &  83.3 ± 13.3 \\
               \hspace{2em}Functional groups$^r$ ($n=30$) &   0.0 ± 0.0 &   0.0 ± 0.0 &   3.3 ± 6.4 &    16.7 ± 13.3 &  30.0 ± 16.4 \\
       \hspace{2em}Benzene$^c$ ($n=20$) &   0.0 ± 0.0 &   5.0 ± 9.6 & 80.0 ± 17.5 &     95.0 ± 9.6 &  100.0 ± 0.0 \\
          \hspace{2em}Benzene$^r$ ($n=20$) & 20.0 ± 17.5 &   0.0 ± 0.0 & 40.0 ± 21.5 &    75.0 ± 19.0 &  90.0 ± 13.1 \\
      \hspace{2em}Pyridine$^c$ ($n=20$) &   0.0 ± 0.0 &   0.0 ± 0.0 & 80.0 ± 17.5 &     95.0 ± 9.6 &  100.0 ± 0.0 \\
         \hspace{2em}Pyridine$^r$ ($n=20$) &   5.0 ± 9.6 & 10.0 ± 13.1 & 55.0 ± 21.8 &    65.0 ± 20.9 &   95.0 ± 9.6 \\
     \hspace{2em}Isoxazole$^c$ ($n=20$) &   0.0 ± 0.0 &   0.0 ± 0.0 & 20.0 ± 17.5 &    70.0 ± 20.1 &  80.0 ± 17.5 \\
        \hspace{2em}Isoxazole$^r$ ($n=20$) &   0.0 ± 0.0 &   5.0 ± 9.6 & 30.0 ± 20.1 &    65.0 ± 20.9 &  100.0 ± 0.0 \\
    \hspace{2em}Napthalene$^c$ ($n=20$) &   0.0 ± 0.0 &   5.0 ± 9.6 & 20.0 ± 17.5 &    20.0 ± 17.5 &  40.0 ± 21.5 \\
       \hspace{2em}Napthalene$^r$ ($n=20$) &   0.0 ± 0.0 &   0.0 ± 0.0 &   0.0 ± 0.0 &      0.0 ± 0.0 &  30.0 ± 20.1 \\
     \hspace{2em}Quinoline$^c$ ($n=20$) &   0.0 ± 0.0 &   0.0 ± 0.0 & 10.0 ± 13.1 &    20.0 ± 17.5 &  25.0 ± 19.0 \\
        \hspace{2em}Quinoline$^r$ ($n=20$) &   0.0 ± 0.0 &   0.0 ± 0.0 &   5.0 ± 9.6 &      5.0 ± 9.6 &  30.0 ± 20.1 \\
\bottomrule
\end{tabular}
}
\begin{flushleft}
\item $^c$ Canonical SMILES
\item $^r$ Random SMILES
\end{flushleft}
\end{table}

\begin{table}[H]
\centering
\caption{Reasoning tokens used by each model for additional questions. Token counts for GPT‑4o and GPT‑4o-tags are the output token counts, whereas the values listed for the o3-mini models are specifically the reasoning token counts. Stated ± is standard deviation of values.}
\label{tab:additional_reasoning_token_data}
\resizebox{\linewidth}{!}{
\begin{tabular}{llllll}
\toprule
                          Question Category &  GPT‑4o & GPT‑4o-tags & o3-mini-low & o3-mini-medium & o3-mini-high \\
\midrule
      SMILES to IUPAC\\ \hspace{2em}Functional groups$^c$ ($n=30$) & 40 ± 10 &      46 ± 8 &  1692 ± 574 &    6306 ± 1510 & 19644 ± 5081 \\
               \hspace{2em}Functional groups$^r$ ($n=30$) &  41 ± 9 &      45 ± 7 &  1363 ± 524 &    8514 ± 2223 & 28685 ± 8949 \\
       \hspace{2em}Benzene$^c$ ($n=20$) &  12 ± 2 &      18 ± 2 &  1290 ± 454 &    4406 ± 1383 & 12442 ± 3283 \\
          \hspace{2em}Benzene$^r$ ($n=20$) &  15 ± 2 &      21 ± 2 &  2019 ± 791 &    6342 ± 2074 & 18195 ± 9475 \\
      \hspace{2em}Pyridine$^c$ ($n=20$) &  12 ± 2 &      18 ± 2 &  1088 ± 239 &     2858 ± 823 &  9203 ± 3257 \\
         \hspace{2em}Pyridine$^r$ ($n=20$) &  16 ± 1 &      22 ± 1 &  1651 ± 748 &    4794 ± 2289 & 14608 ± 6809 \\
     \hspace{2em}Isoxazole$^c$ ($n=20$) &  15 ± 4 &      22 ± 5 &  1869 ± 581 &    4227 ± 1594 & 12534 ± 3984 \\
        \hspace{2em}Isoxazole$^r$ ($n=20$) &  19 ± 3 &      26 ± 4 &  2000 ± 670 &    6006 ± 2602 & 14390 ± 5260 \\
    \hspace{2em}Napthalene$^c$ ($n=20$) &  14 ± 2 &      21 ± 3 &  1645 ± 679 &    8794 ± 4078 & 27635 ± 9101 \\
       \hspace{2em}Napthalene$^r$ ($n=20$) &  16 ± 2 &      22 ± 2 &  1536 ± 480 &    8691 ± 1980 & 35296 ± 6780 \\
     \hspace{2em}Quinoline$^c$ ($n=20$) &  14 ± 2 &      20 ± 3 &  1898 ± 562 &   10160 ± 3579 & 29261 ± 9943 \\
        \hspace{2em}Quinoline$^r$ ($n=20$) &  16 ± 2 &      22 ± 2 &  1888 ± 693 &   10202 ± 3862 & 33571 ± 6700 \\
\bottomrule
\end{tabular}
}
\begin{flushleft}
\item $^c$ Canonical SMILES
\item $^r$ Random SMILES
\end{flushleft}
\end{table}

\clearpage
\section{Reasoning excerpts}\label{app:reasoning_excerpts}
\begin{paddedtable}[H]
\caption{Excerpt from o3-mini-high of an example carbon counting question.}
\label{tab:transcript_counting_carbon}
\begin{tabular}{|>{\columncolor{myLightGray}}m{0.1\textwidth}|m{0.9\textwidth}|}
\hline
Prompt & \begin{minipage}[t]{\linewidth}
How many carbon atoms are in the molecule:\\

C(C)Oc1ccc(cc1Cl)-c1nn(cc1C=C(C\#N)C(NCC)=O)-c1ccccc1\\

Give your answer as an integer. Do not write any comments.
\end{minipage} \\
\hline
Question visualization & \includegraphics[width=0.4\linewidth]{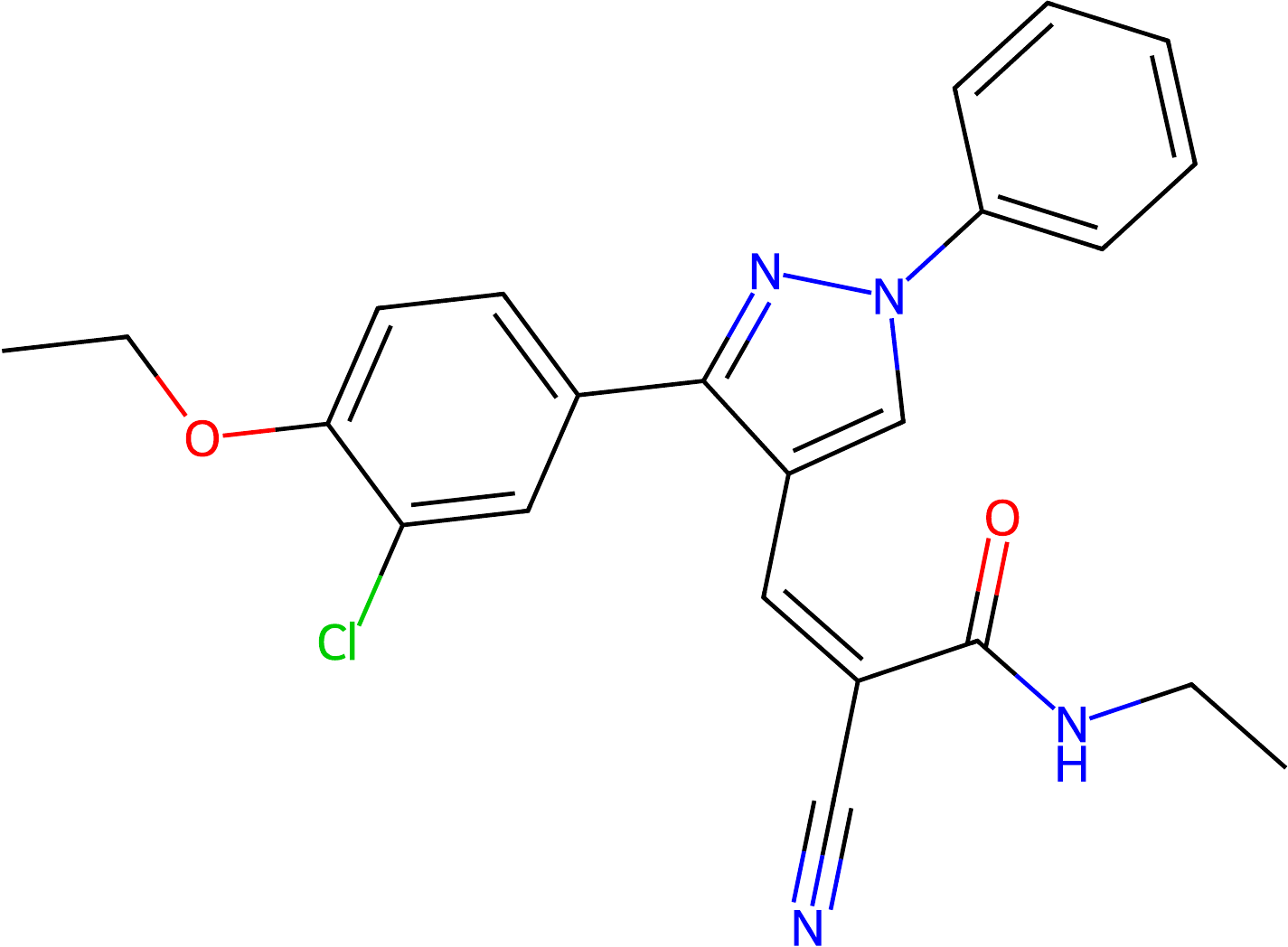} \\
\hline
Reasoning excerpt & \begin{minipage}[t]{\linewidth}
[...]
Let me break it down: "C(C)" refers to a carbon with a methyl group attached to it. This is likely a central carbon being bonded to another methyl group.\\

I'll need to count all the carbon atoms in this structure—terminal carbons, those in attachments, and those in the benzene rings, etc. It’s quite a detailed structure [...]\\

The "C(C)" part is ambiguous—it could mean a methyl branch, potentially forming "CH2-CH3," but I think it’s actually "CH(CH3),” attached to "O," making “CH(CH3)O" with 2 carbons.\\

Next, "c1ccc(cc1Cl)" represents a benzene ring with a chlorine (6 carbons total). The fragment "-c1nn(cc1C=C(C\#N)C(NCC)=O)" seems to be a substituted heterocycle, likely a triazole, though it’s tricky to identify without more context.\\

This molecule has multiple parts that need careful analysis to count all carbon atoms.\\

The molecule consists of three parts:
\begin{itemize}
\item Fragment 1: "C(C)O" with 2 carbons and "c1ccc(cc1Cl)" contributing 6 carbs, totaling 8 carbon atoms.
\item Fragment 3: "c1ccccc1" is a benzene ring contributing 6 more carbon atoms.
\item Fragment 2 is tricky—its structure suggests a heterocyclic ring, possibly a pyrazole or triazole ring, with a branching structure starting with "c1". After careful parsing, I think the branch "C=C(C\#N)C(NCC)=O" isn’t part of the ring but attaches afterward. That helps clarify the ring’s structure.
\end{itemize}

The molecule can be broken into three segments:
\begin{enumerate}
\item "C(C)O c1ccc(cc1Cl)" gives 8 carbon atoms (2 from "C(C)O" and 6 from the benzene ring).
\item "c1nn(cc1C=C(C\#N)C(NCC)=O)" likely contains a 5-membered heterocyclic ring (3 carbons) and a branch "C=C(C\#N)C(NCC)=O" contributing 6 more carbons. So, segment 2 gives 9 carbon atoms.
\item "c1ccccc1" gives 6 carbon atoms (again, a benzene ring).
\end{enumerate}
Summing these, the molecule contains 23 carbon atoms in total.
\end{minipage} \\
\hline
Model Answer &  23 \Large{\checkmark} \\
\hline
\end{tabular}
\end{paddedtable}

\begin{paddedtable}[H]
\caption{Excerpt from o3-mini-high of an example ring counting question. The highlighted text shows that the model recognizes the solution only requires analyzing the ring closure numbers, as opposed to parsing the entire SMILES string.}
\label{tab:transcript_counting_rings}
\begin{tabular}{|>{\columncolor{myLightGray}}m{0.1\textwidth}|m{0.9\textwidth}|}
\hline
Prompt & \begin{minipage}[t]{\linewidth}
How many rings are in the molecule:\\

s1c2n(cc1)cc(n2)C(N1C(c2ccccn2)c2c(CC1)c1ccccc1[nH]2)=O\\

Give your answer as an integer. Do not write any comments.\\
\end{minipage} \\
\hline
Question visualization & \includegraphics[width=0.4\linewidth]{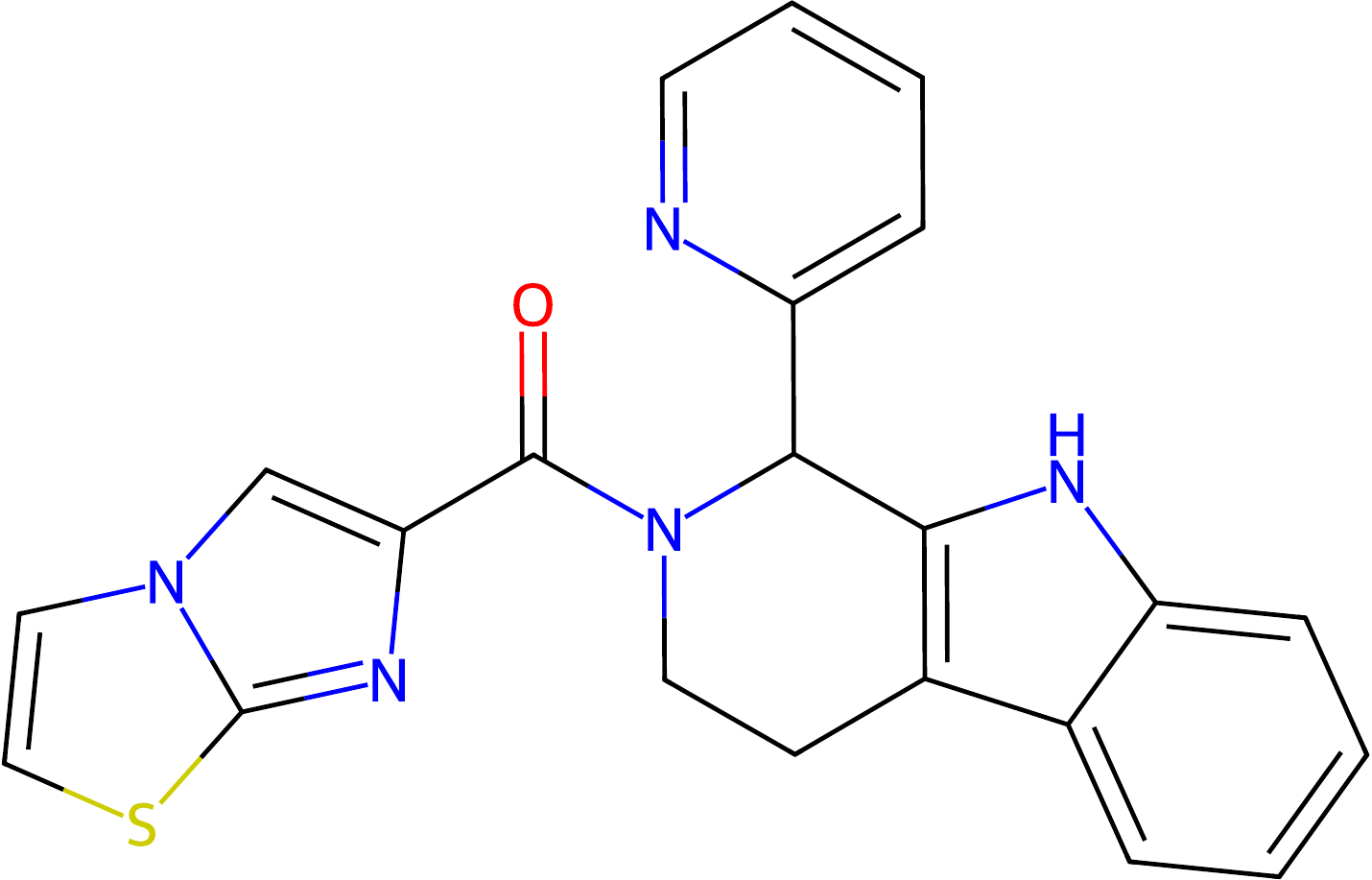} \\
\hline
Reasoning excerpt & \begin{minipage}[t]{\linewidth}
Looking at the molecule's SMILES string:\\ "s1c2n(cc1)cc(n2)C(N1C(c2ccccn2)c2c(CC1)c1ccccc1[nH]2)=O", I need to figure out how many rings are present.\\

I’ll manually parse this, focusing on the ring closure digits like 1 and 2.\\

It seems there are rings marked by those digits, so I need to count them carefully now to find the total.\\

Let’s see... it looks like multiple rings are involved with this structure.\\

The given SMILES string has repeated ring closure digits that represent different rings. It uses digits like "1" and "2" to indicate ring closures.\\

For example, "s1" connects sulfur to a ring with closure 1. Then "c2" opens ring 2. Later, we see "n(cc1)" which means ring 1, and "cc(n2)" means ring 2.\\

When analyzing, \hl{we only need to count the distinct rings based on these digit closures}, meaning there are only two unique rings here: ring 1 and ring 2.\\

In the SMILES string, the digit "1" is used multiple times but it refers to different rings depending on the context. Initially, "s1" opens ring 1, and later "cc1" closes it. Similarly, "n(cc1)" reaffirms ring closure 1, while "cc(n2)" becomes part of ring 2. Later, "N1" initiates a new ring closure, and we see "c2" opening new rings. I need to track each digit carefully, making sure that each ring closure is counted distinctly and correctly.\\
\end{minipage} \\
\hline
Model Answer & 6 \Large{\checkmark} \\
\hline
\end{tabular}
\end{paddedtable}

\begin{paddedtable}[H]
\caption{Excerpt from o3-mini-high of an example shortest path question using canonical SMILES. Highlighted text suggest the model parsed the SMILES string to a graph represents and then performed a graph search to find the shortest path.}
\label{tab:transcript_shortest_path}
\begin{tabular}{|>{\columncolor{myLightGray}}m{0.1\textwidth}|m{0.9\textwidth}|}
\hline
Prompt & \begin{minipage}[t]{\linewidth}
Determine the number of bonds along the shortest path connecting the two dummy atoms (denoted by ’*’). Count each bond equally, including those directly attached to the dummy atoms.\\

*c1cccc2c1sc1c(C)nn(C(C(=O)NCc3ccc(F)cc3)C(*)C)c(=O)c12\\

Give your answer as an integer. Do not write any comments.
\end{minipage} \\

\hline
Question visualization & \includegraphics[width=1\linewidth]{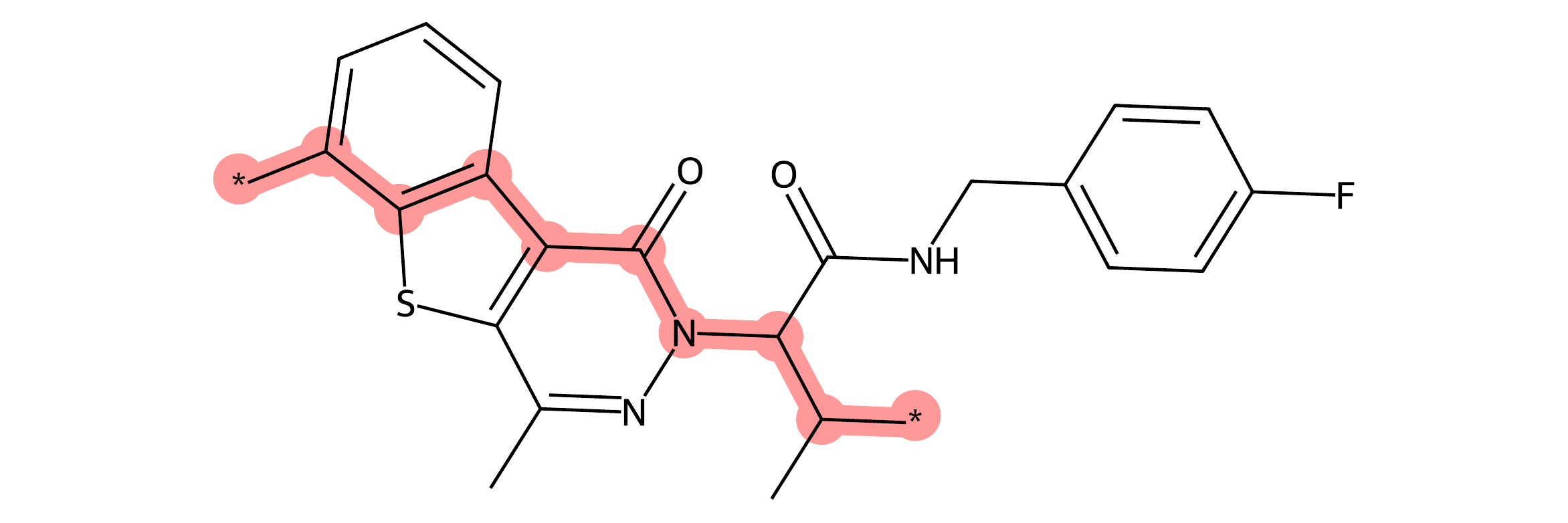} \\

\hline
Reasoning excerpt & \begin{minipage}[t]{\linewidth}
[...]\\

The first thing is recognizing that '*' is a dummy atom. Then we observe an aromatic ring structure with numbered ring closures. I'm labeling atoms starting with the dummy atom as A0. The structure shows connections between various ring atoms, and I’m working out the ring closures to determine the path between the dummy atoms. Let's break this down carefully.\\

The SMILES string is a bit tricky, so let's break it down to understand the bonding. We start with a dummy atom "*" connected to an aromatic carbon labeled "c1". From there, the aromatic ring forms, and we encounter subsequent atoms like sulfur and nitrogen, which are part of an extended structure.\\

The sequence proceeds with a branch, which includes a substituent chain "C(C(=O)NC...)". The tricky bit is parsing the bond structure in this branch, especially the "*" dummy atom, which marks a substituent. I’ll proceed step by step but need to keep track of the ring closing and branch orientations.\\

Okay, let's focus on the structure and clarify the branching. After identifying dummy atom 1 as attached to an aromatic carbon, we see a sulfur atom following further aromatic carbons. There’s a second dummy atom (*), attached later in the chain, at a spot where it connects with a branch. The branch contains a substituent chain, which includes carbonyl and nitrogen atoms, ending in an aromatic ring. This complex branching doesn't affect the connectivity between the dummy atoms, which are part of a polycyclic structure. I’ll keep track of node connections.\\

\hl{To find the shortest path between the two dummy atoms, we need to trace the connections through the molecular graph. While A1 (dummy1) connects through the main chain, B2 (dummy2) is attached to a branch. After examining multiple potential routes, I can list several possibilities. The path D1 -> A2 -> A7 -> A8 -> A9 -> A10 -> A11 -> A12 -> B1 -> B2 -> D2 is one contender, but another slightly shorter path (with 9 bonds) emerges if I navigate via ring closures.}\\

\end{minipage} \\
\hline
Answer & 9 \Large{\checkmark} \\
\hline
\end{tabular}
\end{paddedtable}

\begin{paddedtable}[H]
\caption{Excerpt from o3-mini-high of an example atom mapping question. Model shows multiple layers of abstraction to solve this question. Highlighted in the model answer is an error due to mapping the pyridine ring in the wrong direction.}
\label{tab:transcript_atom_mapping}
\begin{tabular}{|>{\columncolor{myLightGray}}m{0.1\textwidth}|m{0.9\textwidth}|}
\hline
Prompt & \begin{minipage}[t]{\linewidth}
You are given two SMILES strings for the same molecule. Atoms are numbered from left to right, with the first atom having index 0. Only heavy atoms are numbered and mapped.\\

Molecule 1: c1cc(ncc1Cl)Sc1cccc(c1C\#N)F\\

Molecule 2: N\#Cc1c(F)cccc1Sc1ccc(cn1)Cl\\

Determine the mapping of atoms from Molecule 1 to Molecule 2. Provide your answer as a list of tuples, where each tuple contains the atom index from Molecule 1 and its corresponding atom index from Molecule 2. Only write the answer. Do not write any comments.\\
\end{minipage} \\

\hline
Question visualization & \includegraphics[width=0.5\linewidth]{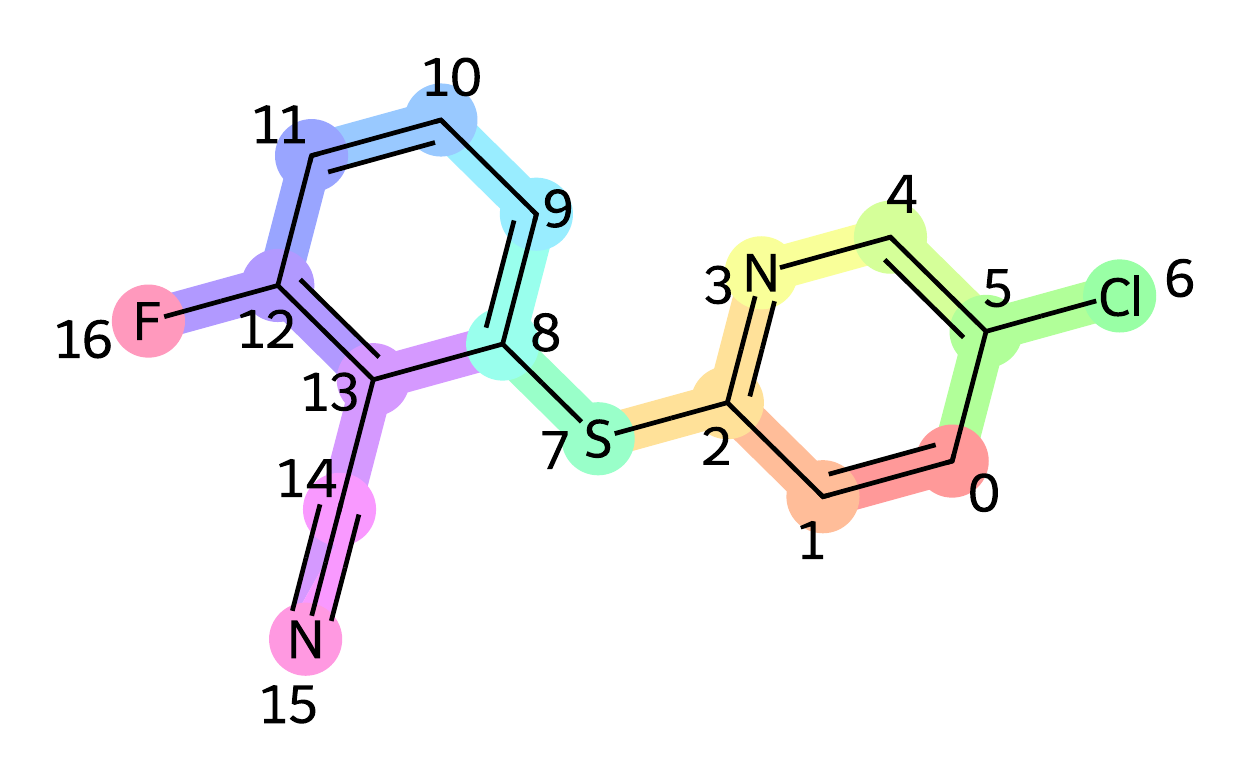}\includegraphics[width=0.5\linewidth]{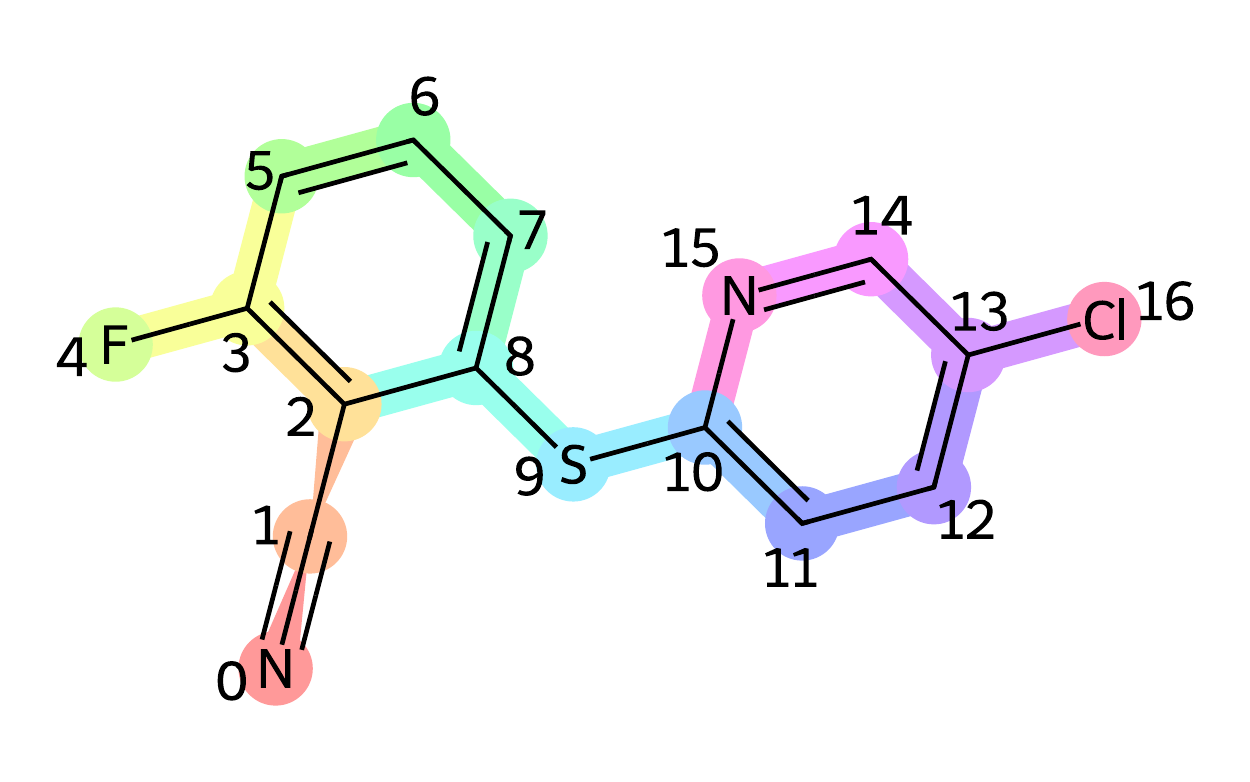} \\

\hline
Reasoning excerpt & \begin{minipage}[t]{\linewidth}
[...] For Molecule 2, I'm numbering the heavy atoms. The string starts with "N\#C," so the first two atoms are nitrogen (0) and carbon (1). Moving left to right, I encounter an aromatic ring starting with "c1"; this is the start of a new ring, making index 2 a "c". After that, "F" is attached to index 3, followed by four more carbons (indices 4 to 8), forming the second aromatic ring. The process continues with an "S" at index 9, then another ring system with substitutions at the end. [...] \\

In Molecule 1, the first benzene ring (Ring A) has Cl, and the second ring (Ring B) has nitrile and F. In Molecule 2, the first ring has nitrile and F, and the second ring has Cl. The S atom joins the rings in both molecules — in Molecule 1, it connects at Ring A (Cl ring) and in Molecule 2, it connects at Ring A (nitrile and F). So the nitrile-F ring in Molecule 1 corresponds to the nitrile-F ring in Molecule 2, and the Cl-laden rings swap places. [...] \\

In Molecule 1, the first ring (Ring 1) has Cl at position M1-5, and S is connected to M1-2, forming a bond with the second ring (Ring 2), where F and nitrile are located at M1-16 and M1-13, respectively. In Molecule 2, the Cl ring (Ring B) is connected to S (M2-9) at M2-10, while the nitrile-F ring (Ring A) has the nitrile at M2-2 and F at M2-3 [...] \\

We need to map the rings and their substituents between Molecule 1 and Molecule 2 carefully, using the positioning of S, F, and nitrile groups. In Molecule 1, Ring 1 (with Cl) and Ring 2 (with F and nitrile) have specific relationships; S connects both in particular positions. For Molecule 2, similar assignments occur for Ring B (with Cl) and Ring A (with nitrile and F), while S still bridges both rings at corresponding positions. [...] 

\end{minipage} \\
\hline
Model Answer & [\hl{(0, 14)}, \hl{(1, 15)}, (2, 10), \hl{(3, 11)}, \hl{(4, 12)}, (5, 13), (6, 16), (7, 9), (8, 8), (9, 7), (10, 6), (11, 5), (12, 3), (13, 2), (14, 1), (15, 0), (16, 4)] \\
\hline
Expected Answer & [(0, 12), (1, 11), (2, 10), (3, 15), (4, 14), (5, 13), (6, 16), (7, 9), (8, 8), (9, 7), (10, 6), (11, 5), (12, 3), (13, 2), (14, 1), (15, 0), (16, 4)] \\
\hline
\end{tabular}
\end{paddedtable}

\begin{paddedtable}[H]
\caption{Reasoning excerpt from o3-mini-high of an example IUPAC naming question.}
\label{tab:transcript_iupac_zinc}
\begin{tabular}{|>{\columncolor{myLightGray}}m{0.1\textwidth}|m{0.9\textwidth}|}
\hline
Prompt & \begin{minipage}[t]{\linewidth}
Write the IUPAC name of this molecule:\\

CCc1ccc(-c2nc(C)c(C(=O)Nc3ccc(NC(C)=O)cc3)s2)cc1\\

Write the IUPAC name only. Do not write any comments.\\
\end{minipage} \\

\hline
Reasoning excerpt & \begin{minipage}[t]{\linewidth}
The user has provided a SMILES string:\\ "CCc1ccc(-c2nc(C)c(C(=O)Nc3ccc(NC(C)=O)cc3)s2)cc1." Let's break it down. The "CC" represents an ethyl group. It then links to a benzene ring (c1ccc...cc1), with a complicated substituent: "-c2nc(C)c(C(=O)Nc3ccc(NC(C)=O)cc3)s2". This looks like a compound with a benzene ring attached to a substituent containing nitrogen, an amide group, and more complex groups. [...]\\

Okay, so I now see that "CCc1ccc(...)" gives an ethyl group bonded to a benzene ring at position 1. Then at another position, there's that complicated group "-c2nc(C)c(...)," which is a heterocyclic thiazole-like ring, with nitrogen, sulfur, and carbon atoms.\\

So the ring is likely a thiazole (1,3-thiazole) with a substitution pattern. The entire structure has two aromatic systems: an ethylphenyl and a thiazole-linked complex. It's tricky, but I’m breaking it down. Let me double-check the naming specifics!\\

Okay, the molecule involves a thiazole ring: it's a 5-membered heterocycle with sulfur, nitrogen, and carbon atoms. The substituents are on carbon atoms 3 and 4. The benzene ring (substituted with ethyl) connects to the heterocycle at position 1. This structure seems to be para-substituted with the ethyl group at one position of the benzene and the thiazole group at the opposing position. [...]\\

Okay, this structure has a 5-membered thiazole ring attached to a phenyl group through a carbamoyl group. The substituents on the thiazole are 4-ethylphenyl at position 2, methyl at position 4, and an N-(4-acetamido)phenyl carbamoyl group at position 5. The ethyl group is positioned para to the heterocycle connection, and the acetamido-group substituted phenyl is attached to the nitrogen of the carbamoyl group. Overall, it looks like the name is: "5-[N-(4-acetamido)phenyl]carbamoyl-4-methyl-2-(4-ethylphenyl)thiazole."\\
\end{minipage} \\
\hline
Model Answer & 5-[N-(4-acetamidophenyl)carbamoyl]-4-methyl-2-(4-ethylphenyl)thiazole \Large{\checkmark} \\
\hline
Visualized molecule & \vspace{0.5em}\includegraphics[width=0.6\linewidth]{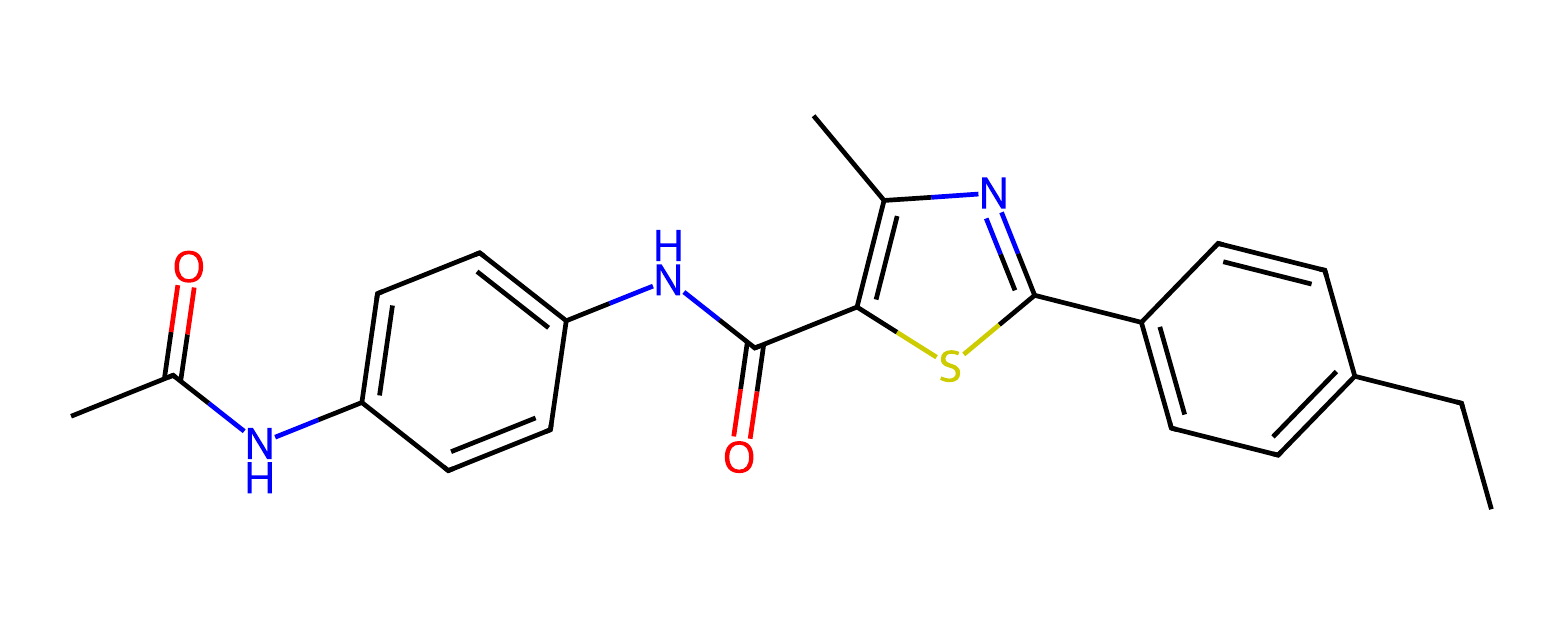}\vspace{0.5em}\\
\hline
\end{tabular}
\end{paddedtable}

\begin{paddedtable}[H]
\caption{Excerpt from o3-mini-high of an example SAR question. The reasoning excerpts show the extraction of R groups from the SMILES string and mapping of the question to a system of equations.}
\label{tab:transcript_SAR}
\begin{tabular}{|>{\columncolor{myLightGray}}m{0.1\textwidth}|m{0.9\textwidth}|}
\hline
Prompt & \begin{minipage}[t]{\linewidth}
Determine the score of the unknown molecule.\\

Index SMILES Score\\
0 c12c(Br)ccc(Br)c1c(F)co2 478\\
1 c12c(Br)ccc(F)c1c(Cl)co2 279\\
2 c12c(Cl)ccc(Br)c1c(Cl)co2 505\\
3 c12c(F)ccc(Br)c1c(Cl)co2 634\\
4 c12c(Br)ccc(Br)c1c(Br)co2 522\\
5 c12c(F)ccc(F)c1c(Br)co2 539\\
6 c12c(Br)ccc(Cl)c1c(F)co2 346\\

Unknown molecule: c12c(F)ccc(Cl)c1c(Br)co2\\

Write your answer as an integer only. Do not write any comments.

\end{minipage} \\
\hline
Reasoning excerpt & \begin{minipage}[t]{\linewidth}
I’m given an unknown molecule with the SMILES string "c12c(F)ccc(Cl)c1c(Br)co2" and its possible score. The known molecules share similar structures with different substituent positions.\\

It seems like the scoring could involve comparing the positions of X, Y, Z — the halogen substituents (F, Cl, Br) — and finding patterns or interpolation between them. I'll consider the substitutions and try to match or approximate the score based on these patterns. Let's break down the possible patterns in the table.\\

Okay, based on the provided molecules, I can use the SMILES string pattern "c12c(X)ccc(Y)c1c(Z)co2" and assign positions as A, B, and C. For each index in the table, A, B, C correspond to various halogen substitutions at specific positions. The unknown molecule has X = F, Y = Cl, Z = Br. I’m thinking that maybe the score results from some combination of these values, potentially in a linear model summing contributions from those substituents. Let me set up some equations to explore the relationships further.\\

I'm working through a system where the score depends on the effects of various substitutions at positions A, B, and C. Based on 7 known examples, I wrote equations for each molecule. Now, I want to determine the unknown score by considering the differences between equations that only differ in one substituent. By subtracting these equations, I get relationships like ($b_Br$ - $b_F$) = 199, and ($c_Br$ - $c_F$) = 44. Now I need to use these relationships to estimate the unknown score for the molecule with A=F, B=Cl, and C=Br.
\end{minipage} \\
\hline
Answer & 576 \Large{\checkmark} \\
\hline
\end{tabular}
\end{paddedtable}

\begin{paddedtable}[H]
\caption{Reasoning excerpt from o3-mini-high for an example product of reaction question using a click reaction. The model quickly recognizes the intended reaction from the functional groups and copper ion. The model initially incorrectly names the alkyne, however this is checked and resolved in the following statement. The model uses a combination of IUPAC names, SMILES strings, structural formulae, and atom indexes to reason to the answer.}
\label{tab:transcript_click_reaction}
\begin{tabular}{|>{\columncolor{myLightGray}}m{0.1\textwidth}|m{0.9\textwidth}|}
\hline
Prompt & \begin{minipage}[t]{\linewidth}
Write the product of the reaction as a SMILES string.\\

[Cu+].[N-]=[N+]=Nc1ccccc1 + C\#CCCCl\\

Only write the SMILES string. Do not write any comments.\\
\end{minipage} \\

\hline
Reasoning excerpt & \begin{minipage}[t]{\linewidth}
[...] The structure "[N-]=[N+]=N" represents an azide group. In the reaction, the azide is attached to a phenyl group ("c1ccccc1") alongside copper(I), forming a copper-azide species. This is relevant to a copper-catalyzed azide-alkyne cycloaddition (Huisgen cycloaddition, or click chemistry). The second reactant, "C\#CCCCl", appears to be "1-chloro-4-pentyn-3", a terminal alkyne with a chlorine substituent. [...]\\

The notation "C\#CCCCl" represents a structure with a triple bond between the first two carbons. [...] The correct name appears to be "3-chlorobut-1-yne" [...]\\

In the copper-catalyzed azide-alkyne cycloaddition (CuAAC), the product is typically a 1,4-disubstituted 1,2,3-triazole. The azide’s substituent (phenyl) attaches to the N1 position of the triazole, while the alkyne’s substituent (the C-CH2CH2Cl part) attaches at the C4 position. The alkyne undergoes a transformation in the reaction, creating a saturated CH2CH2Cl group instead of maintaining the triple bond. The final product should be 1-phenyl-4-(2-chloroethyl)-1,2,3-triazole. [...]\\

The product, 1-(phenyl)-4-(2-chloroethyl)-1,2,3-triazole, has a substituent at N1 (phenyl) and at C4 (CH2CH2Cl). For the right SMILES, we use "c1ccccc1[n]2nnc(CCCl)c2", where:
\begin{itemize}
\item "c1ccccc1" is the phenyl group,
\item "[n]2" denotes the nitrogen at position 1,
\item "nnc" represents N2, N3, and C4 atoms, and
\item "CCCl" is the 2-chloroethyl group attached to C4.
\end{itemize}
This structure reflects the 1,4-disubstitution pattern common to CuAAC reactions.
\end{minipage} \\
\hline
Model Answer & c1ccccc1[n]2nnc(CCCl)c2 \Large{\checkmark} \\
\hline
Visualized reaction & \vspace{0.5em}\includegraphics[width=1\linewidth]{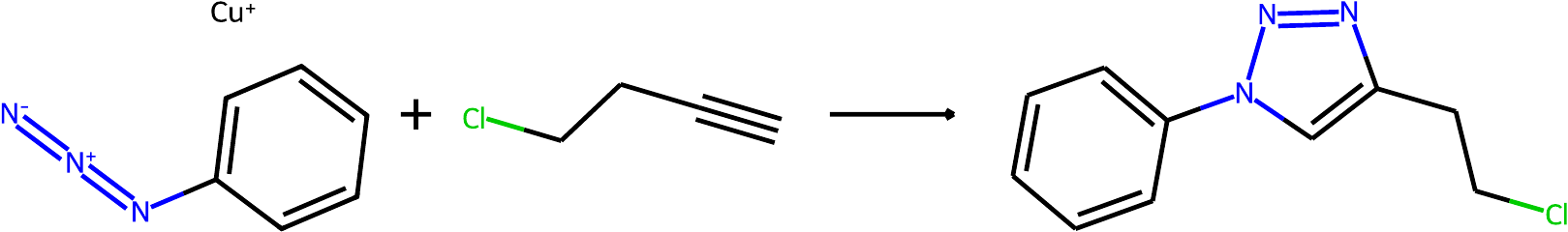}\vspace{0.5em}\\
\hline
\end{tabular}
\end{paddedtable}

\begin{paddedtable}[H]
\caption{Excerpt from o3-mini-high of an example NMR elucidation question using a small molecule. The model shows identification of substructures using both $^{1}$H and $^{13}$C NMR spectra and the molecular formula.}
\label{tab:transcript_nmr_small}
\begin{tabular}{|>{\columncolor{myLightGray}}m{0.1\textwidth}|m{0.9\textwidth}|}
\hline
Prompt & \begin{minipage}[t]{\linewidth}
Write the SMILES string of the molecule consistent with this data:\\

Formula: C6H14N2O\\

1H NMR: $\delta$ 1.40-1.64 (4H, 1.47 (tt, J = 7.4, 7.3 Hz), 1.58 (quint, J = 7.4 Hz)), 1.86 (3H, s), 2.63 (2H, t, J = 7.3 Hz), 3.17 (2H, t, J = 7.4 Hz).\\

13C NMR: $\delta$ 23.1 (1C, s), 26.9 (1C, s), 27.9 (1C, s), 40.0 (1C, s), 41.0 (1C, s), 179.7 (1C, s).\\

Only write the SMILES string. Do not write stereochemistry. Do not write any comments.\\
\end{minipage} \\

\hline
Reasoning excerpt & \begin{minipage}[t]{\linewidth}
The user wants the SMILES string for a molecule with the formula C6H14N2O based on NMR data. The presence of a carbonyl group at 179.7 ppm suggests it's likely an acid. The molecule has two nitrogen atoms and one oxygen — which directs me toward options like a carboxylic acid or amide. However, guanidine doesn’t fit the formula, so I’ll need to think about other possibilities that meet this structure. Let's narrow this down further!\\

Looking at the 1H NMR data, the signals suggest two different CH2 groups (1.40-1.64), a methyl singlet (1.86), and two CH2 groups next to a carbonyl or nitrogen (2.63 and 3.17). The carbonyl from a potential acetamide shows at 179.7 ppm.\\

The hydrogen count is odd, though. The integration adds up to 11 hydrogens, but the formula requires 14. I’m wondering if there are exchangeable NH protons that weren't observed in the NMR — that's probably where the extra hydrogens are.\\

After reconsidering the molecule's structure and the NMR data, the formula C6H14N2O fits with N-acetyl-1,4-diaminobutane. This structure has 6 carbon atoms, 14 hydrogen atoms, 2 nitrogen atoms, and 1 oxygen atom, matching the formula exactly.\\

For the NMR signals: the singlet at 1.86 ppm corresponds to the acetyl group, while the two triplets from 2.63 and 3.17 ppm match CH2 groups adjacent to nitrogen and a carbonyl. This molecule seems to fit well!\\

\end{minipage} \\
\hline
Model Answer & CC(=O)NCCCCN \Large{\checkmark} \\

\hline
Visualized molecule & \includegraphics[width=0.5\linewidth]{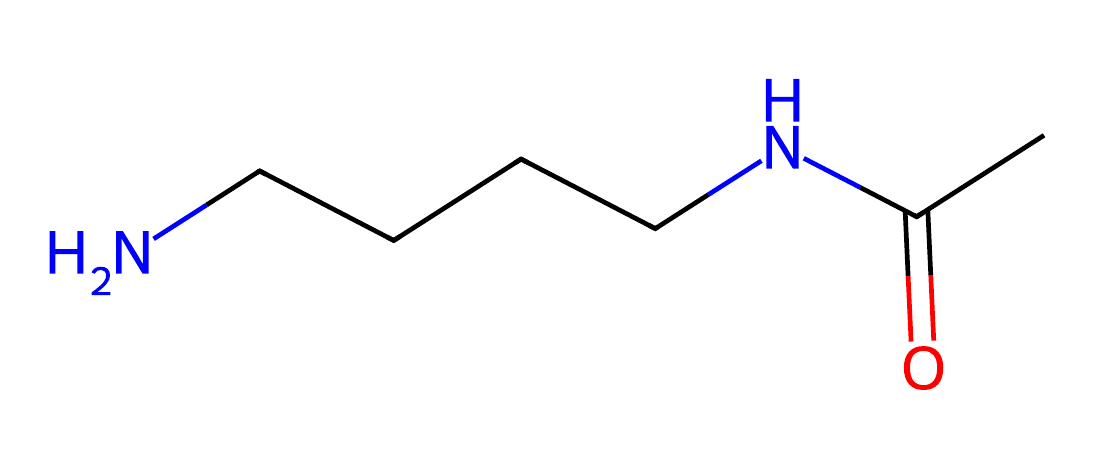}\\
\hline
\end{tabular}
\end{paddedtable}

\begin{paddedtable}[th]
\caption{Example full prompt for a 2D NMR elucidation question. Spectra were simulated using the nmrium NMR prediction web tool (www.nmrium.org/predict) \cite{Patiny2024}}
\label{tab:full_zinc_prompt}
\begin{tabular}{|>{\columncolor{myLightGray}}m{0.1\textwidth}|m{0.9\textwidth}|}
\hline
Prompt & \begin{minipage}[t]{\linewidth}
{\small
Write the SMILES string of the molecule consistent with this data.\\

\textbf{Formula}: C19H19N3O3\\

\textbf{1H NMR}: $\delta$ 7.63 (ddd, J = 7.88, 1.41, 0.51 Hz, 1H), 7.6 (ddd, J = 8.14, 7.52, 1.41 Hz, 1H), 7.48 (dddd, J = 8.16, 1.47, 1.24, 0.54 Hz, 2H), 7.39 (ddd, J = 8.14, 1.39, 0.51 Hz, 1H), 7.34 (ddd, J = 7.88, 7.52, 1.39 Hz, 1H), 7.27 (dddd, J = 8.16, 7.76, 1.43, 0.54 Hz, 2H), 7.07 (tt, J = 7.76, 1.24 Hz, 1H), 4.66 (s, 2H), 3.85 (t, J = 6.97 Hz, 2H), 1.71 (tq, J = 6.97, 6.61 Hz, 2H), 0.96 (t, J = 6.61 Hz, 3H).\\

\textbf{13C NMR}: $\delta$ 169.1 (1C, s), 161.4 (1C, s), 152.5 (1C, s), 140.5 (1C, s), 138.07 (1C, s), 135.15 (1C, s), 129 (1C, s), 128.9 (2C, s), 126.75 (1C, s), 123.58 (1C, s), 120.2 (2C, s), 115.6 (1C, s), 115.2 (1C, s), 47.35 (1C, s), 42.6 (1C, s), 21 (1C, s), 11.15 (1C, s).\\

\textbf{COSY} ($\delta$H, $\delta$H): (7.6, 7.39), (7.6, 7.34), (7.63, 7.34), (7.07, 7.27), (7.48, 7.27), (3.85, 1.71), (1.71, 0.96).\\

\textbf{HSQC} ($\delta$H, $\delta$C): (7.6, 135.15), (7.63, 129), (7.27, 128.9), (7.34, 126.75), (7.07, 123.58), (7.48, 120.2), (7.39, 115.2), (4.66, 47.35), (3.85, 42.6), (1.71, 21), (0.96, 11.15).\\

\textbf{HMBC} ($\delta$H, $\delta$C): (4.66, 169.1), (7.63, 161.4), (3.85, 161.4), (3.85, 152.5), (4.66, 152.5), (4.66, 140.5), (7.6, 140.5), (7.63, 140.5), (7.39, 140.5), (7.48, 138.07), (7.27, 138.07), (7.63, 135.15), (7.39, 135.15), (7.34, 135.15), (7.6, 129), (7.34, 129), (7.07, 128.9), (7.48, 128.9), (7.27, 128.9), (7.6, 126.75), (7.63, 126.75), (7.39, 126.75), (7.48, 123.58), (7.27, 123.58), (7.07, 120.2), (7.48, 120.2), (7.27, 120.2), (7.63, 115.6), (7.39, 115.6), (7.34, 115.6), (7.6, 115.2), (7.34, 115.2), (1.71, 42.6), (0.96, 42.6), (3.85, 21), (0.96, 21), (3.85, 11.15), (1.71, 11.15).\\

Only write the SMILES string. Do not write stereochemistry. Do not write any comments.\\
}
\end{minipage} \\
\hline
Visualized molecule & {\vspace{0.5em}\includegraphics[width=0.6\linewidth]{figures/large_2d_mol.png}}\\
\hline
\end{tabular}
\end{paddedtable}

\clearpage
\begin{figure}[th]
  \includegraphics[width=\textwidth]{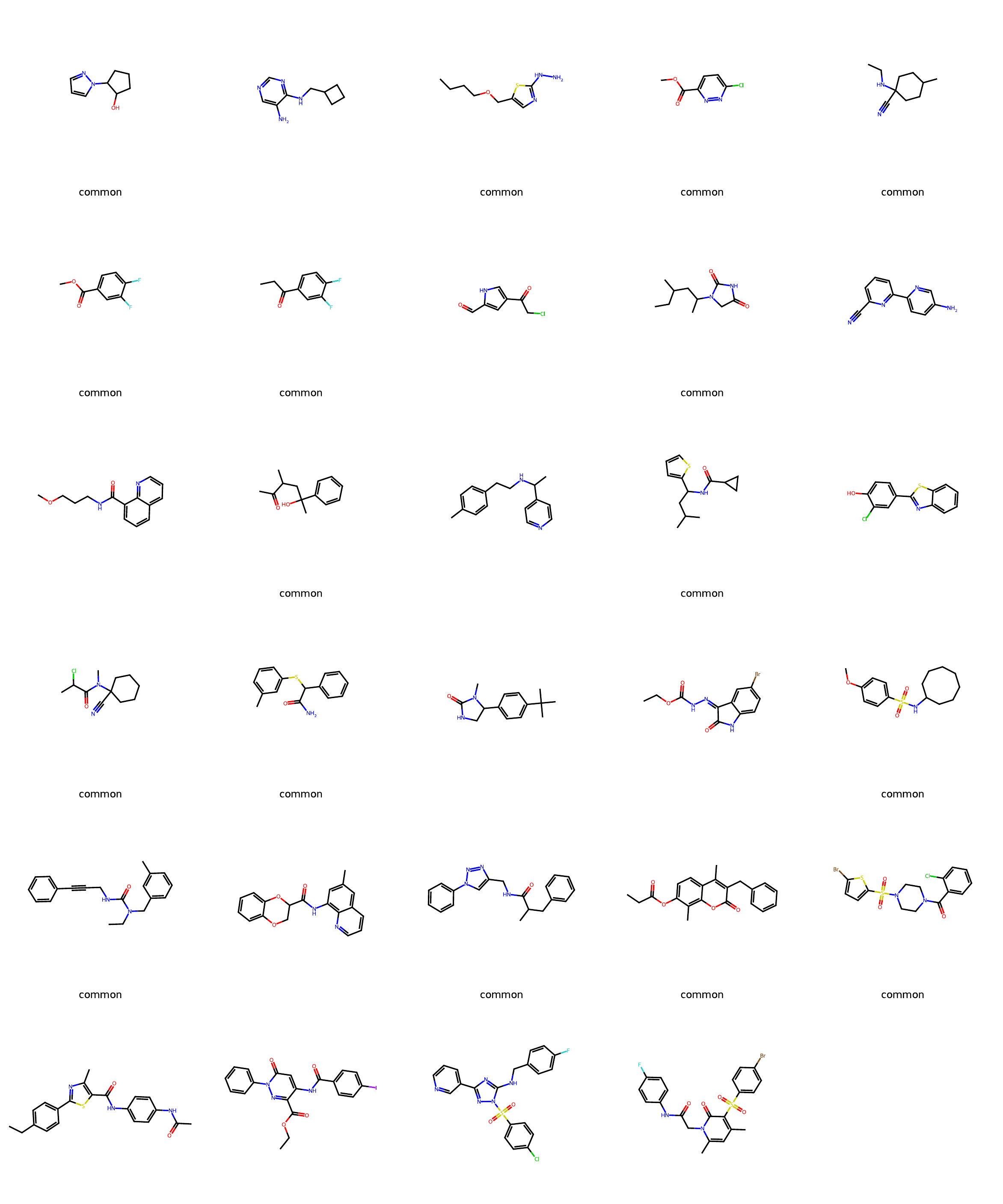}
  \caption{ZINC molecules successfully named by o3-mini-high (from canonical SMILES). ``Common'' signifies this molecule was also successfully named from randomized SMILES.}
  \label{fig:zinc_iupac_canonical_smiles}
\end{figure}

\clearpage
\begin{figure}[H]
  \includegraphics[width=\textwidth]{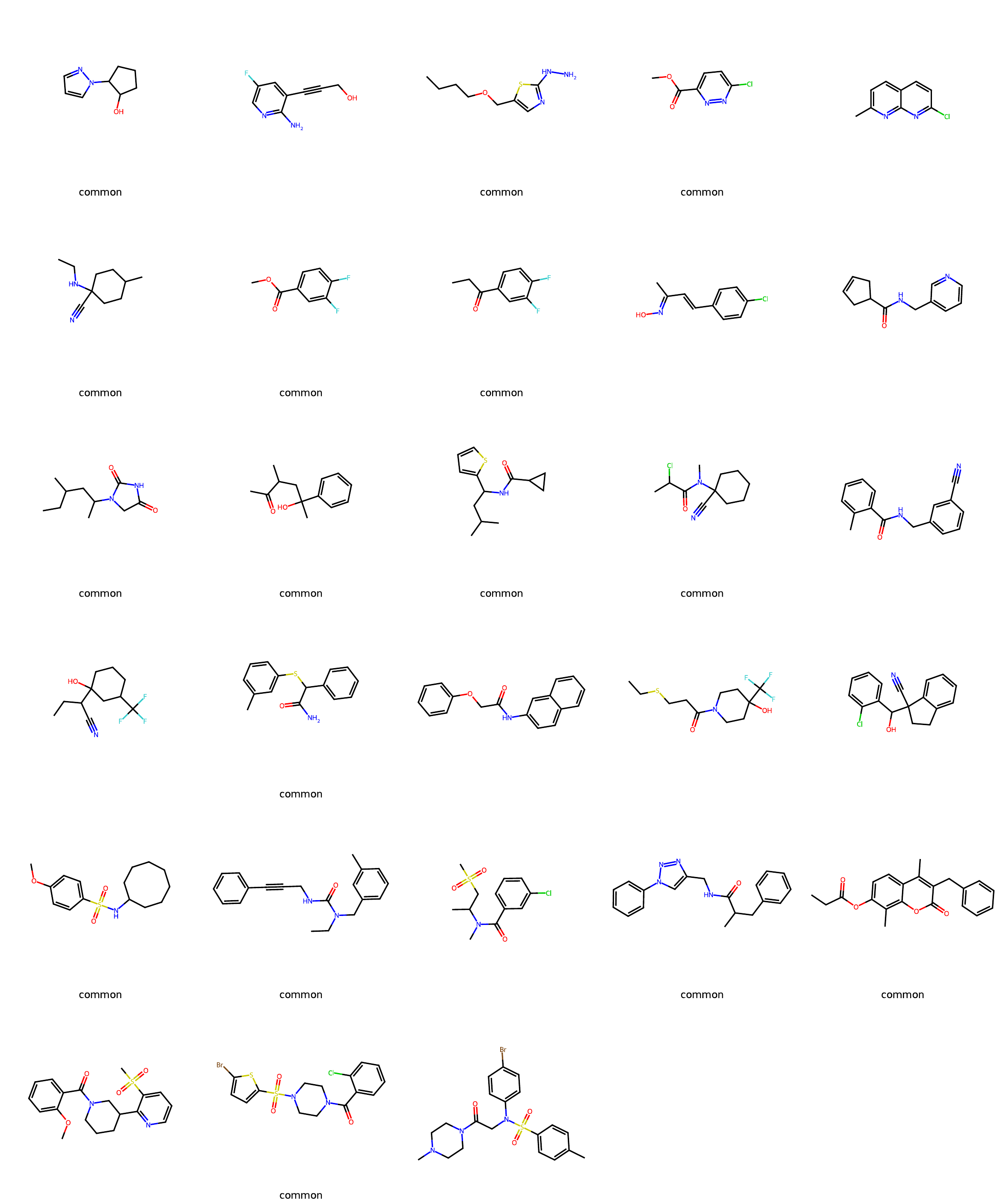}
  \caption{ZINC molecules successfully named by o3-mini-high (from random SMILES). ``Common'' signifies this molecule was also successfully named from canonical SMILES.}
  \label{fig:zinc_iupac_random_smiles}
\end{figure}

\clearpage
\begin{figure}[th]
  \includegraphics[width=\textwidth]{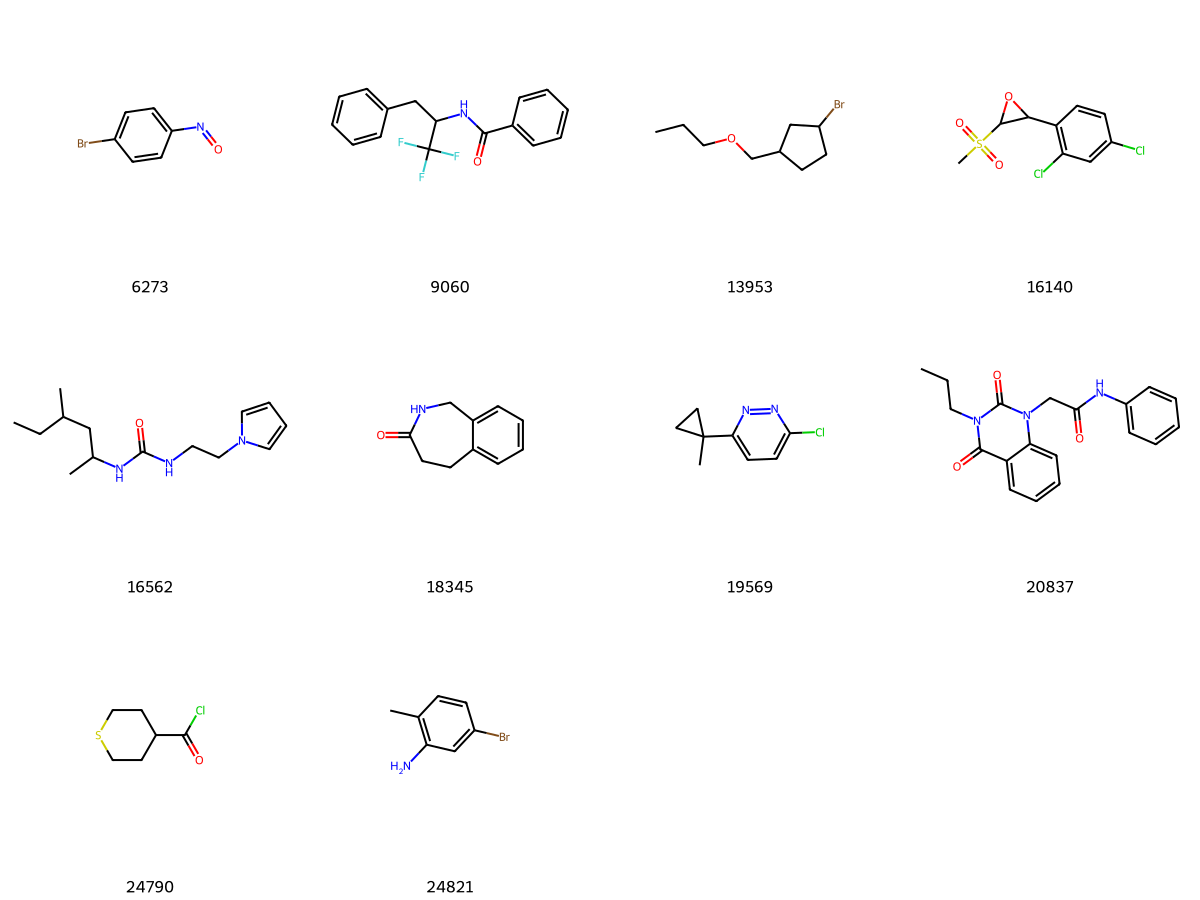}
  \caption{2D NMR structures solved by Gemini Pro 2.5 (reasoning token budget of 32768). Molecules were sampled from ZINC. An example prompt is given in \ref{tab:full_zinc_prompt}. The caption for each molecule is the number of reasoning tokens used by the model to solve the question.}
  \label{fig:gemini_pro_2d_nmr_correct}
\end{figure}

\begin{paddedtable}[H]
\caption{This question is directly from ChemBench \cite{mirza_are_2024} and is not part of our benchmark. Excerpt from o3-mini-medium suggests the question was solved using a process of elimination and does not require a deep understanding of the connectivity of the molecules. Highlighted in the question prompt are features referenced in the reasoning excerpt.}
\label{tab:IUPAC_to_SMILES_MCQ_reasoning}
\begin{tabular}{|>{\columncolor{myLightGray}}m{0.1\textwidth}|m{0.9\textwidth}|}
\hline
Prompt & \begin{minipage}[t]{\linewidth}
What is a valid SMILES for a molecule with the name N’-[(E)-[1-(4-chlorophenyl)-2,5-dimethyl-pyrrol-3-yl]methyleneamino]-N-cyclohexyl-malonamide? \\

A) C\hl{COC(=O)c}1ccc(NC(=O)CSc2nnc(CCc3ccc(C)o3)n2-c2ccccc2)cc1 \\
B) \hl{COc1}ccc(C(=O)O)cc1\hl{S(=O)(=O)N}1CCSCC1 \\
C) C\hl{c1cc}(/C=N/\hl{NC(=O)CC(=O)N}C2CCCCC2)\hl{c}(C)\hl{n1}-c1ccc(Cl)cc1\\
D) O=C1NC(=O)C(N2CCc3ccccc32)=C1c1ccccc1Cl\\
\end{minipage} \\

\hline
Question visualization & \vspace{0.5em}\includegraphics[width=0.9\linewidth]{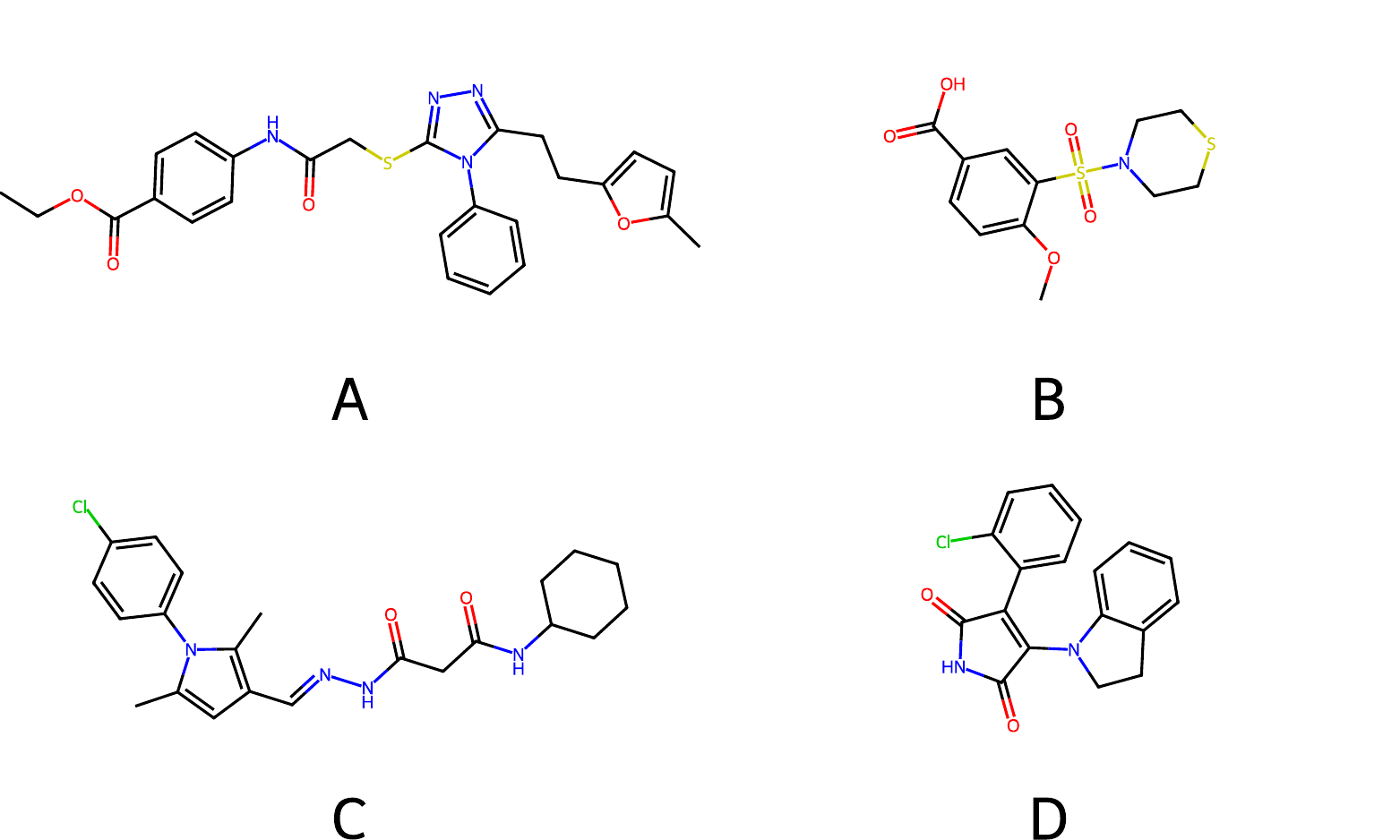} \\

\hline
Reasoning excerpt & \begin{minipage}[t]{\linewidth}
The user is asking for a valid SMILES for a complex molecule. I’ll start by parsing the name: “N’-[(E)-[1-(4-chlorophenyl)-2,5-dimethyl-pyrrol-3-yl]methyleneamino]-N-cyclohexyl-malonamide.” \\

The name suggests a malonamide structure with substitutions. The structure seems to have one nitrogen substituted with cyclohexyl and the other with a specific methyleneamino group. So, the core structure is N-cyclohexyl malonamide, modified by the (E)-substituent. I’ll look at the given options now. \\

The key part of the molecule seems to be the malonamide core, where one nitrogen is substituted with a cyclohexyl group and the other with an (E)-methyleneamino group attached to a pyrrole. Looking at the SMILES options: \\

\hl{Option A is way off, featuring ester and unrelated groups. Option B includes a methoxybenzene and sulfonamide — not matching.} \\

\hl{Option C seems promising, as it contains pyrrole with the correct substituents and malonamide-like structure. The substituent appears to match the described features. I'll go with C.}\\
\end{minipage} \\
\hline
Model Answer & C \Large{\checkmark} \\
\hline
\end{tabular}
\end{paddedtable}

\end{document}